\documentclass[runningheads]{llncs}
\usepackage{eccv}
\usepackage{fontawesome5}
\usepackage{subcaption}
\usepackage{eccvabbrv}
\usepackage{wrapfig}
\usepackage{multirow}

\usepackage{amsmath,amsfonts,bm}

\def\eqref#1{equation~\ref{#1}}

\def\1{\bm{1}}

\DeclareMathAlphabet{\mathsfit}{\encodingdefault}{\sfdefault}{m}{sl}
\SetMathAlphabet{\mathsfit}{bold}{\encodingdefault}{\sfdefault}{bx}{n}

\usepackage{graphicx}
\usepackage{booktabs}
\usepackage{makecell}
\usepackage{mathtools}
\usepackage{array}
\usepackage{marvosym}
 
\usepackage[table]{xcolor} 
\usepackage[accsupp]{axessibility}  
\usepackage[breaklinks,colorlinks,citecolor=eccvblue]{hyperref}
\usepackage{orcidlink}
\definecolor{lightblue}{RGB}{220,235,255}
\newcommand{\method}{GUI-AIMA}

\begin{document}
\title{\texorpdfstring{
GUI-AIMA: \textbf{A}ligning \textbf{I}ntrinsic\\
\textbf{M}ultimodal \textbf{A}ttention with a Context Anchor\\
for GUI Grounding
}{GUI-AIMA: Aligning Intrinsic Multimodal Attention with a Context Anchor for GUI Grounding}}
\titlerunning{GUI-AIMA}
\authorrunning{S. Zhou et al.}
\author{Shijie Zhou\inst{1}\thanks{Work done at UB through University Collaborations.}
Viet Dac Lai\inst{2}
Hao Tan\inst{2}
Jihyung Kil\inst{2}
Wanrong Zhu\inst{2}\\
Changyou Chen\inst{1}
Ruiyi Zhang\inst{2}\thanks{Project lead; work done while at Adobe Research; \textsuperscript{\Letter}Corresponding author.}\textsuperscript{\Letter}}
\institute{University at Buffalo, USA \and Adobe Research, USA\\
\email{\{shijiezh, changyou\}@buffalo.edu, ryzhang.cs@gmail.com}}
\maketitle
\begin{abstract}
Graphical user interface (GUI) grounding is a key capability for computer-use agents, mapping natural-language instructions to actionable regions on the screen. Existing Multimodal Large Language Model (MLLM) approaches typically formulate GUI grounding as a text-based coordinate generation task. However, directly generating precise coordinates from visual inputs is challenging and often data-intensive.
A more intuitive strategy is to first \textit{identify instruction-relevant visual patches and then determine the exact click location within them}.
Motivated by recent observations that general MLLMs exhibit native grounding ability embedded in their attention maps,
we propose \method{},
an attention-based and coordinate-free supervised fine-tuning framework for efficient GUI grounding.
\method{} aligns the intrinsic multimodal attention of MLLMs with patch-wise grounding signals. These signals are calculated adaptively for diverse user instructions by multi-head aggregation on simplified query-visual attention matrices. Besides, its coordinate-free manner can easily integrate a plug-and-play zoom-in stage. \method{}-3B was trained with only 509k samples ($\sim$101k screenshots), demonstrating exceptional data efficiency and verifying that light training can trigger the native grounding capability of MLLMs. It achieves state-of-the-art performance among 3B models, attaining an average accuracy of \textbf{61.5\%} on ScreenSpot-Pro, \textbf{92.1\%} on ScreenSpot-v2, \textbf{68.1\%} on OSWorld-G, \textbf{79.1\%} on MMBench-GUI-L2 and \textbf{60.0\%} on UI-Vision.
Project page: \url{https://github.com/sjz5202/GUI-AIMA}.
\end{abstract}
\section{Introduction}
Graphical User Interface (GUI) agents~\cite{hong2024cogagent,cheng2024seeclick,liang2026anticipatory} have emerged as pivotal tools in automating interactions with digital devices, spanning from mobile applications~\cite{rawles2023androidinthewild,rawles2024androidworld,wang2024mobile,ye2025mobile} to desktop software~\cite{zhang2024ufo,qin2025ui,deng2023mind2web,xie2024osworld,cua2025,zheng2024gpt}. 
GUI grounding is a core capability of GUI agents, requiring the model to map natural-language instructions to specific interface elements on the screen, such as buttons, text fields, and icons~\cite{yang2024aria}. Early GUI grounding methods often rely on structured interface metadata, such as accessibility trees for mobile applications~\cite{zhang2025appagent}. Although such representations provide explicit descriptions of interface elements, they are limited in accessibility, often overly verbose, and may miss critical visual cues such as layout and icons.
These limitations have motivated a shift from metadata-based LLM grounding to screenshot-based multimodal large language models (MLLMs), which can directly reason over visual interfaces.
\begin{wrapfigure}[13]{r}{0.45\textwidth} 
\centering
\includegraphics[width=\linewidth]{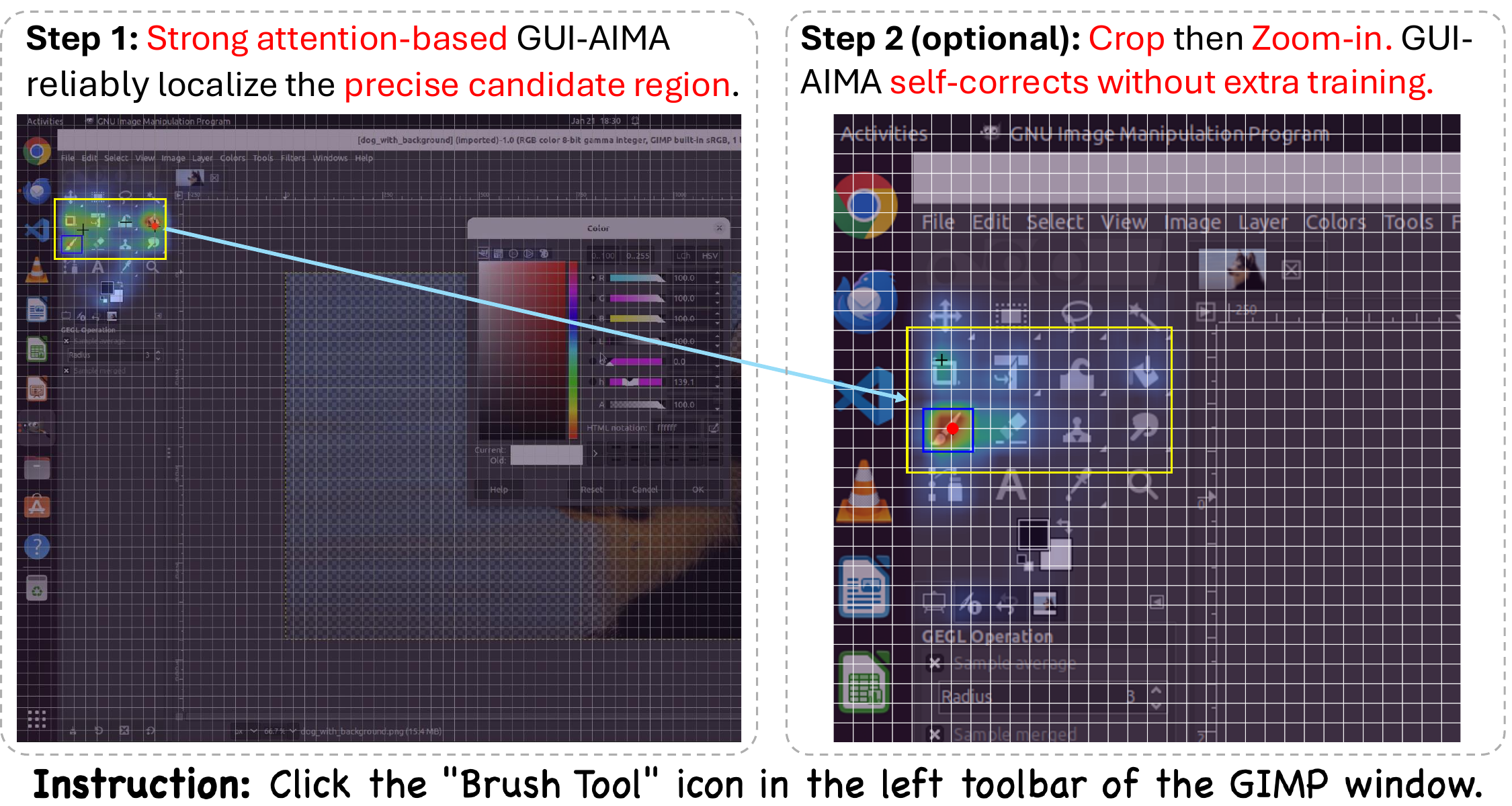}
\caption{An example of \method{} with two-step GUI grounding for high resolution screenshots.}
\label{fig:two_step}
\end{wrapfigure}

However, most existing MLLM-based GUI grounding methods~\cite{gou2024navigating,wu2024atlas,lin2024showui,xu2024aguvis} formulate the task in a coordinate-based manner, i.e., by directly predicting click coordinates as text tokens. This is an indirect formulation, since the model must compress fine-grained visual localization into textual coordinate generation. At the same time, recent evidence suggests that once GUI grounding is performed with MLLMs, useful query-visual grounding signals already emerge in their multimodal attention maps~\cite{xu2024attentiondrivenguigroundingleveraging}. This makes a more natural alternative possible: instead of generating coordinate tokens, one can perform coordinate-free GUI grounding by directly exploiting the MLLM's native attention as patch-level grounding cues, first identifying the relevant visual patches and then determining the specific click center within the selected region.

Recent efforts have begun to explore this direction. TAG~\cite{xu2024attentiondrivenguigroundingleveraging} leverages intrinsic query-visual attention in MLLMs, but aggregates attention maps from all query tokens in a vanilla manner, which is simple but often inaccurate. GUI-Actor~\cite{wu2025gui} also avoids direct coordinate generation, but introduces additional grounding modules instead of utilizing native attention and requires an extra adaptation stage. These limitations leave open an important question: can we directly specialize the \emph{intrinsic} multi-head attention of MLLMs for GUI grounding, without extra modules and without relying on impractical token-wise aggregation?

In this work, we propose \method{}, a coordinate-free GUI grounding framework that directly supervises the MLLM's multi-head self-attention (MHSA)~\cite{vaswani2017attention} with patch-wise grounding signals.
\method{} enables simple and fine-grained aggregation of patch-level attention vectors across all query tokens and all attention heads through two designs. First, we append a learnable $\texttt{<ANCHOR>}$ token after visual and query tokens, and use its attention over visual patches as a surrogate aggregator of query-visual interactions, thereby avoiding explicit token-wise aggregation. Second, we introduce a head-weighting mechanism based on \emph{visual-sink query tokens}: a small subset of query-side tokens selected from hidden-state similarity that reflects a strong query-visual connection. These tokens are used as proxies to identify attention heads that are more likely to encode grounding-relevant cross-modal correspondences, enabling more precise multi-head aggregation while minimally perturbing less grounding-relevant heads.

Built on this coordinate-free attention formulation, \method{} naturally supports practical improvements in both training and inference. We design an overlap- and distance-aware patch labeling that better matches click behavior. Moreover, because \method{} produces grounding from patch-level attention, it naturally enables a two-step zoom-in inference procedure for high-resolution screenshots: the model first localizes a coarse region, and then refines the prediction on a cropped view without additional training, as shown in \cref{fig:two_step}. This helps correct offset errors that commonly arise on dense, high-resolution interfaces.

\method{} is both effective and data-efficient. Under the ablation training setting, it improves ScreenSpot-Pro accuracy by 5.88\% over vanilla attention grounding (43.39\% over 37.51\%). With only one-stage fine-tuning on 101k screenshots without extra modules, \method{}-3B outperforms similarly sized models and remains competitive with substantially larger MLLM-based GUI grounding methods.

Overall, contributions of \method{} are as follows: \textbf{(i)} We propose \method{}, an attention-based, coordinate-free framework that directly leverages the intrinsic attention of MLLMs for GUI grounding, without introducing extra grounding modules; \textbf{(ii)} We develop a simple attention aggregation strategy that replaces impractical vanilla token-wise aggregation and a head-weighting mechanism that emphasizes grounding-relevant attention heads; \textbf{(iii)} We design a patch-wise labeling that better matches click behavior, together with a two-step zoom-in inference procedure for improved high-resolution localization; \textbf{(iv)} With only one-stage fine-tuning on 101k screenshots, \method{}-3B achieves strong data efficiency, outperforming similarly sized models and remaining competitive with substantially larger GUI grounding systems.
\section{Related Work}
\noindent\textbf{Coordinate-based GUI grounding:} The core challenge for GUI agents~\cite{wang2024gui} is grounding: aligning user instructions with the correct actionable elements in the screenshot, due to the semantic inconsistency of layouts and UI elements across diverse GUI environments~\cite{liu2025infigui}. In early attempts, along with the screenshots, extra structured inputs, such as HTML for U-Ground~\cite{gou2024navigating}, or extractions from visual parsing modules, such as element extractions from OmniParser~\cite{Wan_2024_CVPR,yu2025omniparserv2structuredpointsofthoughtunified} and generated captions as contexts in Aria-UI~\cite{yang2024aria}, are fed into MLLMs as the supplementary inputs for a better interface understanding and more precise localization. Later works, such as AGUVIS~\cite{xu2024aguvis}, SeeClick~\cite{cheng2024seeclick} and OS-Atlas~\cite{wu2024atlas}, explore the GUI grounding leveraging MLLMs~\cite{bai2025qwen2,chen2024expanding} with screenshot-only inputs, enabling the end-to-end localization with improved scalability across diverse GUI environments. The grounding approach in these methods is to generate coordinate-based click centers or bounding boxes as text, which is indirect alignment of visual grounding requiring additional efforts, such as scaled GUI corpora~\cite{qin2025ui,chai2024amex} and OCR pretraining~\cite{hong2024cogagent} for connecting coordinates with UI elements. Recent endeavors manage to address this gap from the data-intensive manner of the coordinate-based GUI grounding methods via incorporating GUI-specific grounding reward signals into RL-training~\cite{luo2025gui,lu2025ui,liu2025infigui,tang2025gui} and performing autonomous GUI exploration~\cite{fan2025gui,wu2025gui}. 

\noindent\textbf{Coordinate-free GUI grounding:} Recent works have begun to replace traditional coordinate‑based grounding text with patch‑level attention map predictions, where patches belonging to the correct region receive higher weight. TAG~\cite{xu2024attentiondrivenguigroundingleveraging} first leveraged the multi-head self-attention between GUI instructions and visual tokens in MLLMs for tuning‑free GUI grounding, showcasing the intrinsic potential of MLLMs' attention patterns for GUI tasks. However, the generalization of TAG for novel interfaces is restricted by the backbone and by the heuristics used for selecting text tokens and attention heads. SE-GUI~\cite{yuan2025enhancing} retains a coordinate‑based prediction manner but employs self‑attention to iteratively filter noisy training samples during training. Aside from these attention-based improvements, GUI-Actor~\cite{wu2025gui} introduces an embedding‑based grounding head that derives attention between input visual patch embeddings and the final hidden state of a special \texttt{<ACTOR>} token. Compared with \method{}, GUI-Actor adds an additional module to connect cross-layer visual and \texttt{<ACTOR>} embeddings, requiring an extra warm-up training phase and resulting in much lower training efficiency.
\section{Methodology}
\begin{figure}[t]
    \centering
    \includegraphics[width=0.95\linewidth]{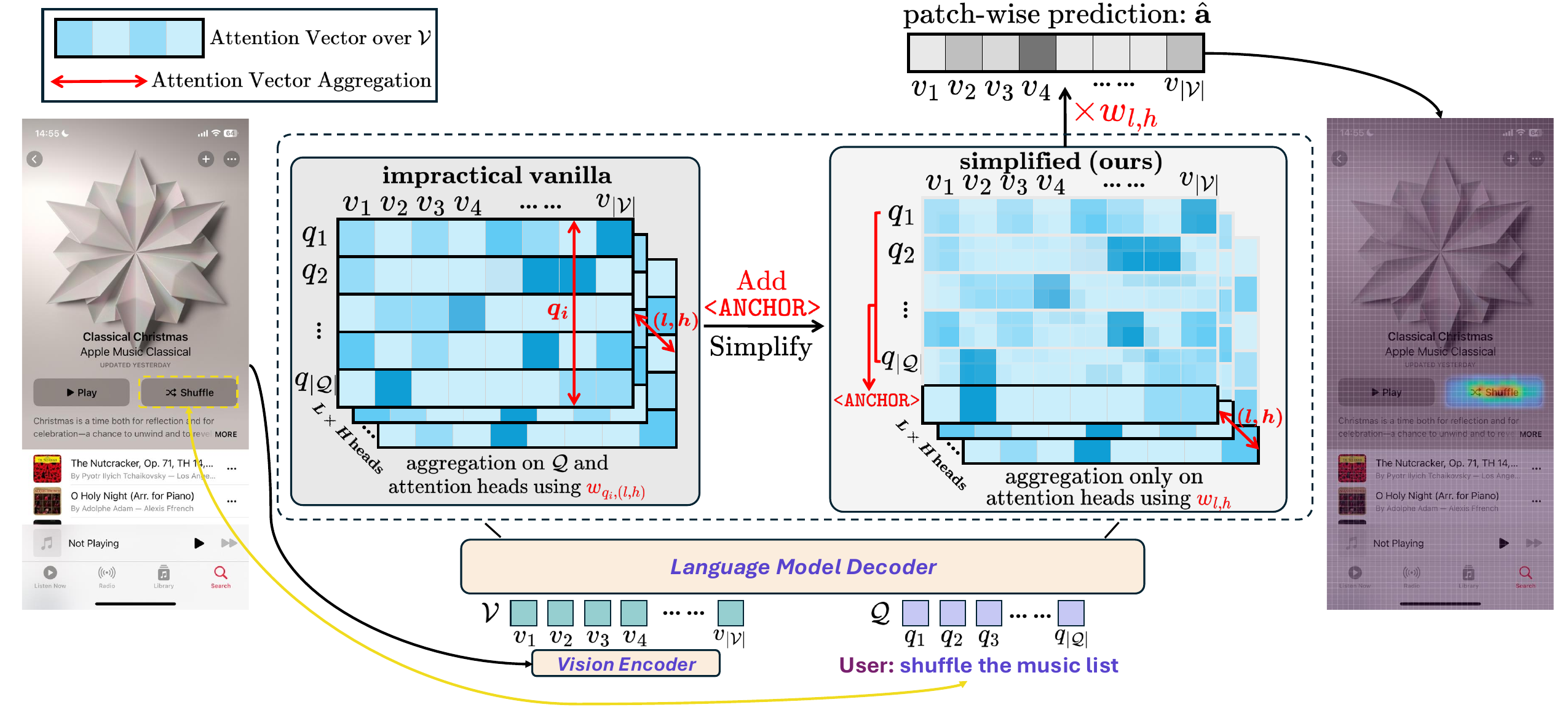}
    \caption{With user query $\mathcal{Q}$, screenshot patches $\mathcal{V}$ and multi-head attentions $\{\mathbf{A}^{l,h}\}_{l\in[L],h\in[H]}$ from the MLLM, the vanilla attention grounding needs additional aggregation between all query tokens' grounding vectors. In our proposed simplified version, a special $\texttt{<ANCHOR>}$ token can learn to implicitly aggregate all query tokens. Then we aggregate grounding vectors of $\texttt{<ANCHOR>}$ token across layers and heads with carefully designed weights to produce the patch-wise predictions.} 
\label{fig:comparison_example}
\end{figure}
\method{} formulates GUI grounding as the attention supervision: it uses the MLLM’s intrinsic multi-head attention maps across layers as grounding indicators, instead of coordinate tokens, and directly supervises these attention maps without introducing additional grounding modules.
As illustrated in \cref{sec:pre}, within each attention head of the MLLM, the \textbf{patch-level attention vector} from a single text token (of the user query) to all visual tokens indicates the relevance of each visual patch to the target grounding region, naturally serving as grounding cues. 
\method{} fully exploits the attention vectors from the entire text sequence and all attention heads in a simple but precise way: adding a special representative token for query-level aggregation (\cref{sec:pre}) and adaptive head-wise reweighting based on multimodal importance (\cref{sec:head_agg}).
\subsection{Intrinsic Attention for GUI Grounding with an Anchor Token} \label{sec:pre}
\noindent\textbf{Preliminaries on Multi-Head Self-Attention (MHSA).}
For each layer $1\leq l \leq L$ of MLLMs, each head $1\leq h \leq H$ computes the attention map $\mathbf{A}^{l,h}$ from preceding layer's embeddings $\mathbf{H}^{l-1}\in\mathbb{R}^{d}$ via $\mathbf{A}^{l,h}=\operatorname{softmax}\bigl(\mathbf{Q}^{l,h} \mathbf{K}^{l,h^{\top}} / \sqrt{d_h}+$\linebreak$\mathbf{M}\bigr)$, where $\mathbf{Q}^{l,h}$ and $\mathbf{K}^{l,h}$ are linear projections of $\mathbf{H}^{l-1}$ with dimension $d_h=d/H$, and $\mathbf{M}$ is the causal mask with $\mathbf{M}_{ij}=0$ for $i\ge j$ and $\mathbf{M}_{ij}=-\infty$ for $i<j$.

\noindent\textbf{Intrinsic Grounding Signals in Attention Maps $\mathbf{A}^{l,h}$.}
Let $\mathcal{V}=[v_1,\ldots,v_{|\mathcal{V}|}]$ and $\mathcal{Q}=[q_1,\ldots,q_{|\mathcal{Q}|}]$ denote the visual patch-token sequence and the user-query text token sequence, respectively. 
Given the attention map $\mathbf{A}^{l,h}$, the \textbf{patch-level attention vector} from query token $q_i$ to all visual patches $\mathcal{V}$ is denoted by $\mathbf{A}^{l,h}_{q_i,\mathcal{V}}$. Its entry $\mathbf{A}^{l,h}_{q_i,v_j}$ measures how strongly patch $v_j$ is associated with token $q_i$, and patches with larger attention values are more likely to contain regions relevant to $q_i$. Therefore, aggregating $\{\mathbf{A}^{l,h}_{q_i,\mathcal{V}}\}_{l,h,i}$ across layers, heads, and query tokens with weight $w_{q_i,(l,h)}$ produces intrinsic grounding cues over all visual patches:
\begin{equation}
\hat{\mathbf{a}}
= \frac{1}{L\cdot H\cdot|\mathcal{Q}|}\sum_{l,h,i}
\, w_{q_i, (l,h)}\,\mathbf{A}^{l,h}_{q_i,\mathcal{V}} \quad\in\mathbb{R}^{|\mathcal{V}|},
\label{eq:aggre_vanilla}
\end{equation}
where $w_{q_i,(l,h)}\ge 0$ and $\sum_{l,h,i} w_{q_i,(l,h)}=1$. 
We refer to \cref{eq:aggre_vanilla} as \textbf{Vanilla Attention Grounding} (``vanilla'' in \cref{fig:comparison_example}).
However, effectively supervising $\hat{\mathbf{a}}$ under this formulation requires a carefully calibrated balance not only across attention heads, but also across $\mathbf{A}^{l,h}_{q_i,\mathcal{V}}$ from different query tokens, making $w_{q_i,(l,h)}$ impractical. 
Moreover, directly imposing supervision on the attention of query tokens may interfere with the pretrained language-vision behaviors~\cite{samaran2021attending}, potentially impairing pre-existing knowledge of MLLM. 

To address these issues, we introduce a special visual anchor token $\texttt{<ANCHOR>}$ and append it \emph{after} the GUI inputs, forming the sequence $[\mathcal{V},\mathcal{Q},\texttt{<ANCHOR>}]$. 
This placement is crucial under causal self-attention: the last token can attend to \emph{all} preceding visual and query tokens, enabling $\texttt{<ANCHOR>}$ to summarize the entire instruction context in its attention distribution. 
As a result, token-wise weighting is avoided by letting $\texttt{<ANCHOR>}$ serve as a representative query token for grounding, while disentangling GUI grounding supervision from the attentions of pre-existing query tokens. 

Concretely, for each layer-head pair $(l,h)$, we use the \textbf{anchored patch-level attention vector} $\mathbf{A}_{\texttt{a},\mathcal{V}}^{l,h}\in\mathbb{R}^{|\mathcal{V}|}$, i.e., the attention from $\texttt{<ANCHOR>}$ to all visual tokens in $\mathcal{V}$, to learns an implicit aggregation of $\{\mathbf{A}^{l,h}_{q_i,\mathcal{V}}\}_{q_i\in\mathcal{Q}}$ over all query tokens. 
With these anchored attentions, we simplify \cref{eq:aggre_vanilla} into a \textbf{multi-head aggregation} that only weights attention heads:
\begin{equation}
\hat{\mathbf{a}}
= \frac{1}{L\cdot H}\sum_{l,h}
\, w_{l,h}\,\mathbf{A}^{l,h}_{\texttt{a},\mathcal{V}} \quad\in\mathbb{R}^{|\mathcal{V}|},
\label{eq:aggre_final}
\end{equation}
where $w_{l,h}\ge 0$ and $\sum_{l,h} w_{l,h}=1$, and we denote the head-weight vector as
$\bm{w}=[\, w_{l,h} : l\in[L],\; h\in[H] \,]\in\mathbb{R}^{1\times L\cdot H}$.
This \textbf{Simplified Attention Grounding} is shown as ``simplified'' in \cref{fig:comparison_example}.

\begin{figure}[t]
    \centering
    \includegraphics[width=0.95\linewidth]{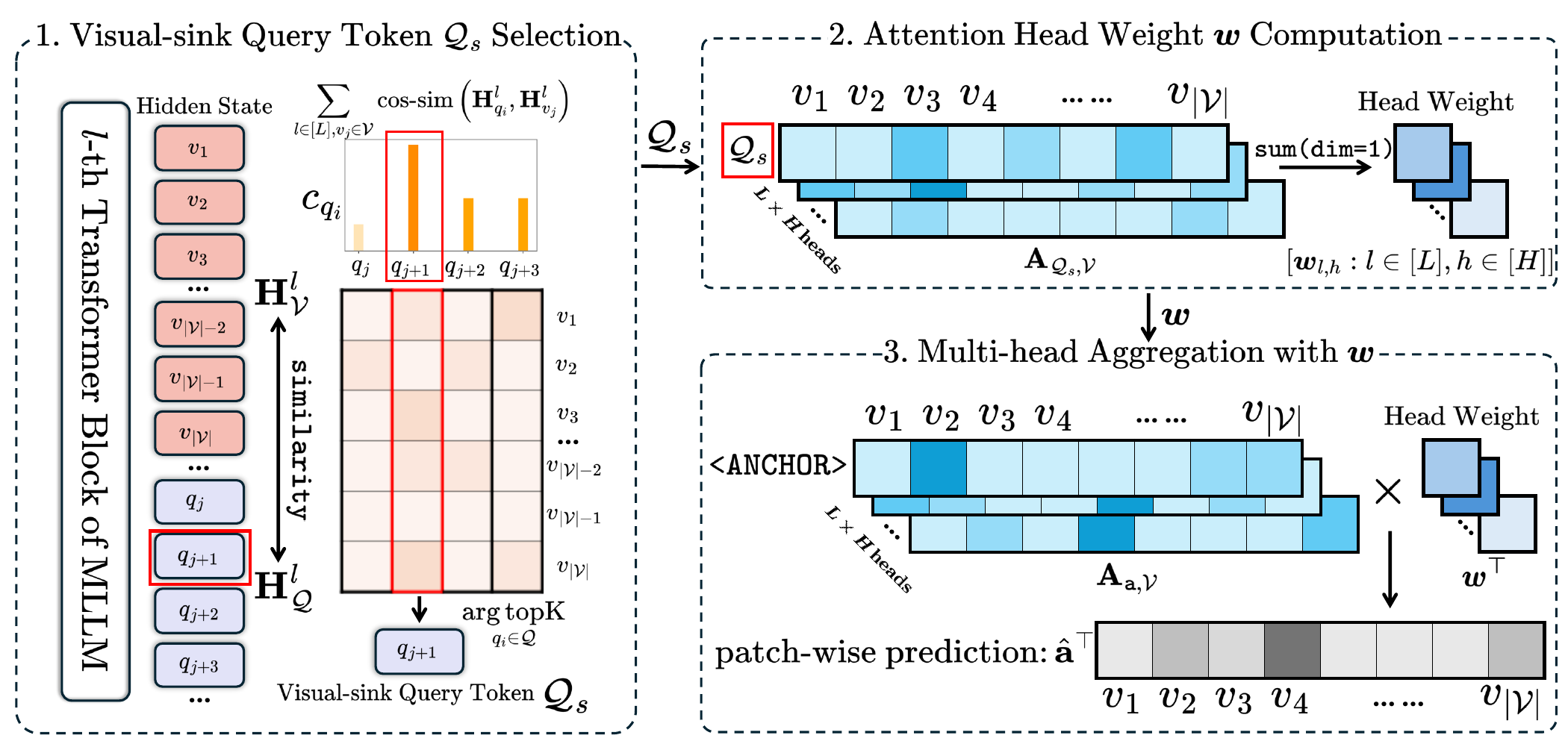}
    \caption{Pipeline of \method{} for patch-wise prediction from the $\texttt{<ANCHOR>}$ token with attention-head weighting: \textbf{(i)} \method{} first selects visual-sink query tokens $\mathcal{Q}_s$ (\cref{eq:find_sink}) via computing hidden state similarities between query tokens and visual patches (\cref{eq:similarity}); \textbf{(ii)} it computes weights $\boldsymbol{w}$ of each attention head (\cref{eq:unified_head_score}) based on selected $\mathcal{Q}_s$; \textbf{(iii)} \method{} aggregates the grounding vectors of $\texttt{<ANCHOR>}$ token across layers and heads (\cref{eq:aggre_final}).} 
    \label{fig:main_fig}
\end{figure}
\subsection{Attention Head Weighting using Visual-sink Query Tokens} \label{sec:head_agg}
In \cref{sec:pre}, we introduce a special $\texttt{<ANCHOR>}$ token to avoid token-wise aggregation over all query words. We now discuss how to determine the head weights $w_{l,h}$ for the multi-head aggregation in \cref{eq:aggre_final}. 
Previous works~\cite{li2023inference,clark2019does} have shown the functional diversity between attention heads across transformer blocks in LLMs. For GUI grounding, we aim to emphasize heads that capture strong query-visual interactions, while down-weighting heads that mainly serve other functions. This selective weighting allows attention supervision to focus on grounding-relevant heads and to minimally perturb neutral heads, helping preserve pretrained capabilities.

\noindent\textbf{An Attention-based View of Head Weighting.}
We measure the grounding relevance of each attention head by the amount of query-to-visual attention mass it carries: heads with stronger query-visual interactions are more likely to encode grounding-relevant correspondences.
For a selected query-side subset $\mathcal{Q}' \subseteq \mathcal{Q}\cup\{\texttt{<ANCHOR>}\}$, we define the head score of attention head $(l,h)$ as the cumulative attention mass from $\mathcal{Q}'$ to all visual tokens, with further normalization:
\begin{equation}
w_{l,h}(\mathcal{Q}')=\sum_{q\in\mathcal{Q}'}\sum_{v_j\in\mathcal{V}} \mathbf{A}^{l,h}_{q,v_j}, \quad w_{l,h}=\frac{\exp \left(w_{l,h}\right)}{\sum_{l'=1}^{L}\sum_{h'=1}^{H}\exp\!\big(w_{l',h'}\big)}
\label{eq:unified_head_score}
\end{equation}

Under this view, different head-weighting strategies can be unified by varying only the query-side token subset used to compute the score. A straightforward choice is to use all query tokens, i.e., $\mathcal{Q}'=\mathcal{Q}$. However, this choice is imprecise, since query tokens contribute unevenly to grounding, and aggregating over all of them can introduce noisy attention mass unrelated to the actual target region. Another simple choice is to use only the anchor token, i.e., $\mathcal{Q}'=\{\texttt{<ANCHOR>}\}$, since $\texttt{<ANCHOR>}$ serves as the representative token for grounding in \cref{eq:aggre_final}. Although simpler, this strategy can be unreliable early in training, because $\texttt{<ANCHOR>}$ is randomly initialized and may not yet serve as a stable context summarizer.

\noindent\textbf{Visual-sink Query Token Selection.}
To estimate grounding-relevant head weights via  \cref{eq:unified_head_score}, we first identify a query-side token subset that captures query-visual interaction in a head-agnostic manner:
for a user query $\mathcal{Q}$, we estimate the visual affinity of each token $q_i$ at layer $l$, denoted as $c_{q_i}^l$, using cosine similarity between its hidden state and all visual patch states (Step~1 in \cref{fig:main_fig}). And We denote the selected top-K tokens via $c_{q_i}^l$ by $\mathcal{Q}_s$:
\begin{equation}
c_{q_i}^l=\sum_{v_j \in \mathcal{V}} \texttt{Sim}(\mathbf{H}^l_{q_i},\mathbf{H}^l_{v_j}).
\label{eq:similarity}
\end{equation}
A larger $c_{q_i}^l$ indicates that the representation of $q_i$ is more strongly aligned with the global visual context at layer $l$.

We use hidden-state similarity instead of directly relying on attention values because grounding-relevant query-visual interactions are often concentrated in only a subset of attention heads, and may therefore not be uniformly observable from raw attention patterns alone~\cite{elhelo2024inferring,olsson2022context,voita2019analyzing}. In contrast, the similarity score computed in \cref{eq:similarity} provides a head-agnostic and more global representation-space signal, making it more stable for identifying tokens that consistently correlate with cross-modal interaction (see the supplementary material for analysis).

As shown in \cref{fig:top_tokens}, the top-ranked token often shows a sink-like behavior: it repeatedly concentrates interaction with the visual context without necessarily
\begin{wrapfigure}[12]{r}{0.57\textwidth}
  \centering
  \includegraphics[width=\linewidth]{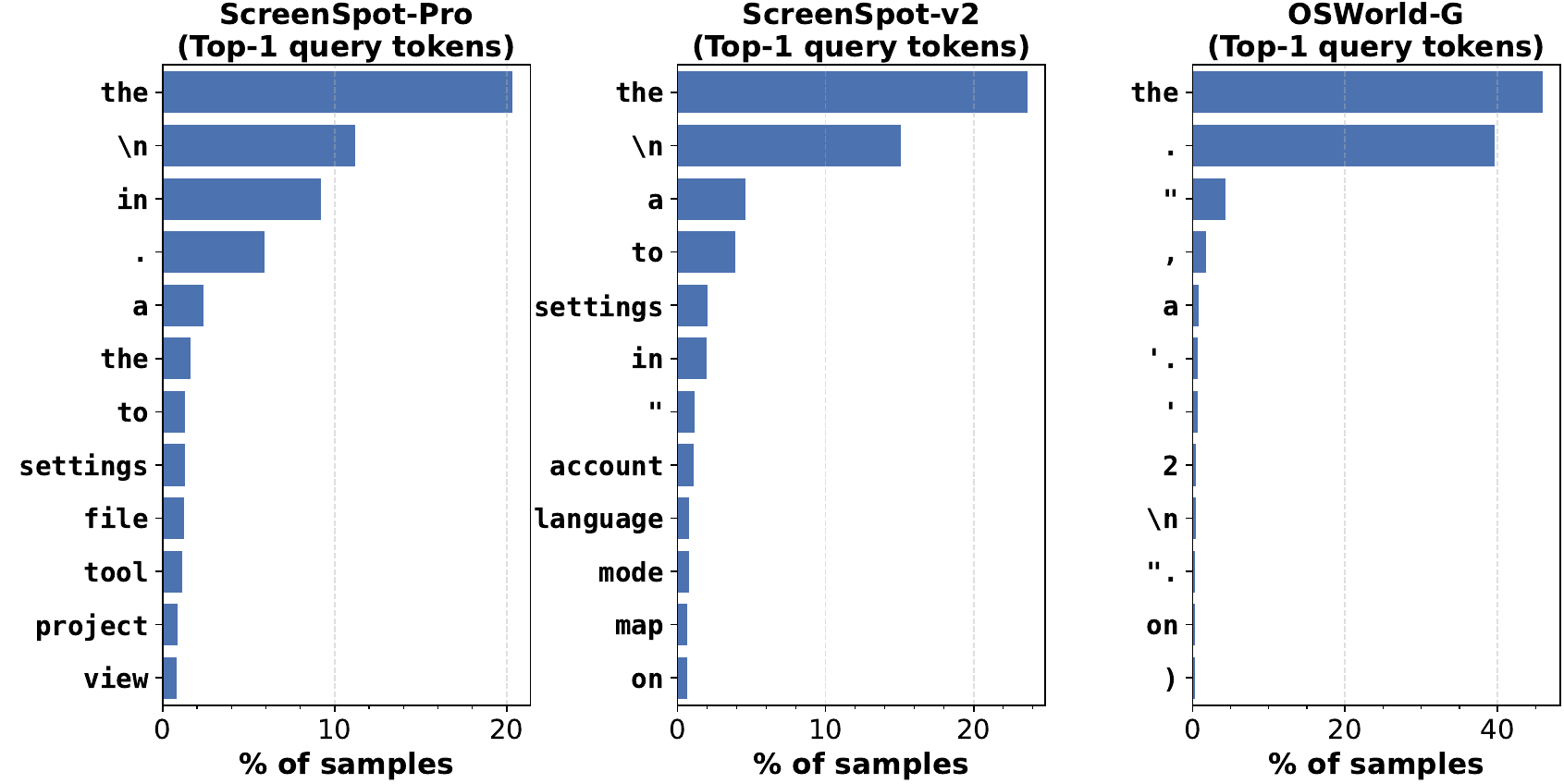}
  \caption{Distribution of top-1 selected visual-sink query tokens under $c_{q_i}$.}
  \label{fig:top_tokens}
\end{wrapfigure}
being the most semantically informative word. We therefore refer to these tokens as \textbf{visual-sink query tokens}, emphasizing their role as \emph{functional proxies} for cross-modal interaction. 
In practice, we find that using a shared token set across all layers provides better stability and performance, instead of layer-wise $\mathcal{Q}_s^l$ for each layer as $\mathcal{Q}_s^l\coloneqq \operatorname*{arg\,topK}_{q_i\in\mathcal{Q}}(c_{q_i}^{\,l})$. Specifically, we adopt a \textbf{global uniform} selection that first aggregates $c_{q_i}^l$ across layers and then selects a shared visual-sink query token set $\mathcal{Q}_s$:
\begin{equation}
\mathcal{Q}_s
\coloneqq \operatorname*{arg\,topK}_{q_i\in\mathcal{Q}}(c_{q_i}),
\quad
c_{q_i}=\sum_{l=1}^{L} c_{q_i}^{\,l}.
\label{eq:find_sink}
\end{equation}

With $\mathcal{Q}_s$, we compute the normalized head weights $w_{l,h}$ using \cref{eq:unified_head_score}. Intuitively, if a head assigns substantial query-to-visual attention mass through these selected tokens, its attention pattern is more consistent with the cross-modal alignment pattern found in representation space. We therefore assign such heads larger weights in \cref{eq:aggre_final} to produce the final patch-level prediction $\hat{\mathbf{a}}$.

\subsection{Training and Inference Design}
\noindent\textbf{Coordinate-free Supervision with Patch-wise Labels.} \label{sec:label}
Since our attention grounding is defined over visual patches rather than pixel coordinates, we convert each ground-truth bounding box $gt^{\mathrm{bbox}}=[x_1,y_1,x_2,y_2]$ into a patch-level label vector $\bm{p}\in\mathbb{R}^{|\mathcal{V}|}$, where $|\mathcal{V}|$ is the number of visual patches. A patch $v_i$ is assigned a positive label only if it overlaps with $gt^{\mathrm{bbox}}$.

To emphasize central patches while softly down-weighting partially overlapped boundary patches, we assign each positive patch a weight based on both its overlap with the ground-truth box and its distance to the box center, following GUI-$\text{G}^2$~\cite{tang2025gui}:
\begin{equation}
    p_{v_i}
    =\textbf{IoU}(v_i,gt^{\mathrm{bbox}})\cdot\mathcal{N}\!\left(\boldsymbol{\mu}_{v_i};\boldsymbol{\mu}_{gt},\boldsymbol{\Sigma}_{gt}\right),
    \label{eq:labeling}
\end{equation}
where $\boldsymbol{\mu}_{v_i}$ and $\boldsymbol{\mu}_{gt}$ denote the centers of patch $v_i$ and $gt^{\mathrm{bbox}}$, respectively, and
$\boldsymbol{\Sigma}_{gt}=\mathrm{diag}\!\left((\sigma_x^{gt})^2,(\sigma_y^{gt})^2\right)$ with
$\sigma_x^{gt}=\alpha(x_2-x_1)$ and $\sigma_y^{gt}=\alpha(y_2-y_1)$. 

The final label vector is normalized as
$\bm{p}=\mathrm{normalize}(\{p_{v_i}\}_{i=1}^{|\mathcal{V}|})$,
which is both overlap-aware and center-biased. We then supervise the predicted patch distribution $\hat{\mathbf{a}}$ from \cref{eq:aggre_final} using the KL divergence:
\[
    \mathcal{L}_{\text{Attn}}=\mathbb{D}_{\mathrm{KL}}(\bm{p}\,\|\,\operatorname{normalize}(\hat{\mathbf{a}})).
\]

\noindent\textbf{Two-step Inference with Zoom-in \faSearchPlus.} \label{sec:zoom}
Under GPU memory constraints, high-resolution GUI screenshots are usually down-sampled, yielding fewer visual patch tokens for the MLLM. The resulting information loss and reduced fine-grained spatial granularity inevitably harm grounding accuracy. 
\method{} provides flexible spatial granularity since its patch-wise grounding. We can easily perform a two-step inference by adding zoom-in \textbf{without extra training}: (1) feed the compressed high-resolution screenshot to predict the approximate location. We use the center of this predicted location to determine a specific area to focus on. (2) we re-run the inference on the newly cropped region, providing a much more accurate result. 
This two-step inference targets to mitigate failure cases on high-resolution screens where the model identifies the right region, but the center prediction is slightly offset from the ground-truth box. Detailed observations and analysis of the zoom-in strategy are provided in \cref{sec:zoomin_analysis}.

\noindent\textbf{Efficient Attention Map Extraction.}
FlashAttention~\cite{dao2022flashattention} avoids materializing full attention maps, while standard eager attention~\cite{vaswani2017attention} can return them but is slower and more memory-intensive. In \method{}, we use FlashAttention for the regular forward pass and eager attention only for the grounding-relevant rows, i.e., the query tokens and $\texttt{<ANCHOR>}$ over visual tokens. This reduces the extraction cost roughly from $O((|\mathcal{V}|+|\mathcal{Q}|+1)^2)$ to $O((|\mathcal{Q}|+1)\cdot|\mathcal{V}|)$. Since $|\mathcal{Q}|\ll |\mathcal{V}|$ in GUI inputs, the additional overhead is negligible, shown in \cref{sec:ablation}.
\section{Experiments}\label{sec:exp}
\subsection{Implementation Details.}
\noindent\textbf{Training Setup.}
We use Qwen2.5-VL-3B-Instruct~\cite{bai2025qwen2} as the MLLM backbone for \method{} and all ablation variants. Following GUI-Actor~\cite{wu2025gui}, we retain the next-token prediction loss for \method{}'s grounding format. \method{}-3B is trained on 8 NVIDIA A100-80G GPUs with a global batch size of 64 and a learning rate of $5\times 10^{-6}$. We set $\alpha=0.8$ for the adaptive deviation control in Gaussian labeling. For visual-sink query token selection, we use the top-1 token under the global uniform criterion in \cref{eq:find_sink}. The patch grid resolution is $28\times 28$ in Qwen2.5-VL base model. \method{}-3B is trained for one epoch with all parameters unfrozen, without extra modules or a warm-up stage.

\noindent\textbf{Training Data and Inference Setup.}
To examine the scalability of \method{} with increased training data, we train two model variants: \method{}-3B-lite is trained on the full training sets of GUIAct~\cite{chen2024guicourse}, AndroidControl~\cite{li2024effects}, and Wave-UI~\cite{waveui}, together with 60k randomly sampled samples from UGround~\cite{gou2024navigating} and 60k from the GTA1 training set~\cite{yang2025gta1guitesttimescaling}, resulting in 259k instructions and approximately 85k images in total. \method{}-3B is then further trained with an additional 250k instructions (509k in total) from GroundCUA~\cite{feizi2025grounding} to better scale to desktop scenarios. For the 45k ablation training set for ablations in \cref{sec:ablation}, we use the first 10k samples from GUIAct, AndroidControl, and Wave-UI, together with the first 15k samples from UGround. For the two-step inference in \cref{sec:zoom}, we set the crop size to $616$ pixels and use a $2\times$ zoom-in ratio in the second pass. Implementation details of the baselines are provided in the supplementary material.

\noindent\textbf{Benchmarks and Metrics.}
We evaluate \method{} on five GUI grounding benchmarks covering a wide range of grounding scenarios: ScreenSpot-Pro~\cite{li2025screenspot}, OSWorld-G~\cite{xie2025scaling} and UI-Vision~\cite{nayak2025ui} (in-domain) for the desktop scenario, Screen-\linebreak -Spot-v2~\cite{wu2024atlas} and MMBench-GUI-L2~\cite{wang2025mmbench} for general evaluations across mobile, desktop, and web domains. 
For grounding accuracy, we follow the standard center-point metric~\cite{lin2024showui}: a prediction is considered correct if the click point falls within the ground-truth bounding box.
\begin{table*}[!t]
    \centering
    \setlength{\tabcolsep}{0.5pt}
    \caption{Performance across various categories in {\bf ScreenSpot-Pro}, reporting Text, Icon, and Average scores. ``-'' indicates unreported results in original papers. Methods with $^*$ are Qwen2.5-VL-based. \faSearchPlus{} denotes two-step inference with zoom-in.}
    \scriptsize
    \resizebox{\textwidth}{!}{
    \begin{tabular}{c l
        c c p{0.4em}
        c c p{0.4em}
        c c p{0.4em}
        c c p{0.4em}
        c c p{0.4em}
        c c p{0.4em}
        c c c}
    \toprule
    & \multirow{2}{*}{\textbf{Model}} 
    & \multicolumn{2}{c}{\textbf{CAD}} 
    & \multicolumn{1}{c}{}
    & \multicolumn{2}{c}{\textbf{Dev}} 
    & \multicolumn{1}{c}{}
    & \multicolumn{2}{c}{\textbf{Creative}} 
    & \multicolumn{1}{c}{}
    & \multicolumn{2}{c}{\textbf{Scientific}} 
    & \multicolumn{1}{c}{}
    & \multicolumn{2}{c}{\textbf{Office}} 
    & \multicolumn{1}{c}{}
    & \multicolumn{2}{c}{\textbf{OS}} 
    & \multicolumn{1}{c}{}
    & \multicolumn{3}{c}{\textbf{Average}} \\
    \cmidrule(lr){3-4}\cmidrule(lr){6-7}\cmidrule(lr){9-10}\cmidrule(lr){12-13}\cmidrule(lr){15-16}\cmidrule(lr){18-19}\cmidrule(lr){21-23}
    & & Text & Icon
    &
    & Text & Icon
    &
    & Text & Icon
    &
    & Text & Icon
    &
    & Text & Icon
    &
    & Text & Icon
    &
    & Text & Icon & \textbf{Avg.} \\
    \midrule
    \multirow{5}{*}{\rotatebox{90}{\bf General}}
    & Claude Computer~\cite{hu2024dawnguiagentpreliminary} & 14.5 & 3.7
    &
    & 22.0 & 3.9
    &
    & 25.9 & 3.4
    &
    & 33.9 & 15.8
    &
    & 30.1 & 16.3
    &
    & 11.0 & 4.5
    &
    & 23.4 & 7.1 & 17.1 \\
    & Qwen2.5-VL-3B~\cite{bai2025qwen2} & 9.1 & 7.3
    &
    & 22.1 & 1.4
    &
    & 26.8 & 2.1
    &
    & 38.2 & 7.3
    &
    & 33.9 & 15.1
    &
    & 10.3 & 1.1
    &
    & 23.6 & 3.8 &  16.1 \\
    & Qwen2.5-VL-7B~\cite{bai2025qwen2} & 16.8 & 1.6
    &
    & 46.8 & 4.1
    &
    & 35.9 & 7.7
    &
    & 49.3 & 7.3
    &
    & 52.5 & 20.8
    &
    & 37.4 & 6.7
    &
    & 38.9 & 7.1 &  26.8 \\
    & Qwen3-VL-8B~\cite{bai2025qwen3} & 56.9 & 10.9
    &
    & 75.3 & 22.8
    &
    & \underline{68.2} & 16.1
    &
    & \underline{78.5} & 32.7
    &
    & \underline{80.8} & 39.6
    &
    & \underline{71.0} & 20.2
    &
    & \bf71.1 & 22.8 &  52.7 \\
    & Qwen3-VL-30B-A3B~\cite{bai2025qwen3} & 51.8 & 15.6
    &
    & \underline{76.0} & 24.8
    &
    & \bf69.2 & 20.3
    &
    & 76.4 & 27.3
    &
    & \underline{80.8} & 37.7
    &
    & \bf75.7 & 38.2
    &
    & \underline{70.6} & 26.3 &  \underline{53.7} \\
    \midrule
    \multirow{9}{*}{\rotatebox{90}{\bf SFT}}
    & OS-Atlas-7B~\cite{wu2024atlas} & 12.2 & 4.7
    &
    & 33.1 & 1.4
    &
    & 28.8 & 2.8
    &
    & 37.5 & 7.3
    &
    & 33.9 & 5.7
    &
    & 27.1 & 4.5
    &
    & 28.1 & 4.0 &  18.9 \\
    & UGround-V1-7B~\cite{gou2024navigating} & 15.8 & 1.2
    &
    & 51.9 & 2.8
    &
    & 47.5 & 9.7
    &
    & 57.6 & 14.5
    &
    & 60.5 & 13.2
    &
    & 38.3 & 7.9
    &
    & 45.2 & 8.1 &  31.1 \\
    & UI-TARS-72B~\cite{qin2025ui} & 18.8 & 12.5
    &
    & 62.9 & 17.2
    &
    & 57.1 & 15.4
    &
    & 64.6 & 20.9
    &
    & 63.3 & 26.4
    &
    & 42.1 & 15.7
    &
    & 50.9 & 17.6 & 38.1 \\
    & \textsc{Jedi}-3B$^*$~\cite{xie2025scaling} & 27.4 & 9.4
    &
    & 61.0 & 13.8
    &
    & 53.5 & 8.4
    &
    & 54.2 & 18.2
    &
    & 64.4 & 32.1
    &
    & 38.3 & 9.0
    &
    & 49.8 & 13.7 &  36.1 \\
    & \textsc{Jedi}-7B$^*$~\cite{xie2025scaling} & 38.0 & 14.1
    &
    & 42.9 & 11.0
    &
    & 50.0 & 11.9
    &
    & 72.9 & 25.5
    &
    & 75.1 & 47.2
    &
    & 33.6 & 16.9
    &
    & 52.6 & 18.2 &  39.5 \\
    & GUI-Actor-3B$^*$~\cite{wu2025gui} & - & -
    &
    & - & -
    &
    & - & -
    &
    & - & -
    &
    & - & -
    &
    & - & -
    &
    & - & - &  42.2 \\
    & GUI-Actor-3B$^*$ + Verifier & - & -
    &
    & - & -
    &
    & - & -
    &
    & - & -
    &
    & - & -
    &
    & - & -
    &
    & - & - & 45.9 \\
    & GUI-Actor-7B$^*$~\cite{wu2025gui} & - & -
    &
    & - & -
    &
    & - & -
    &
    & - & -
    &
    & - & -
    &
    & - & -
    &
    & - & - &  44.6 \\
    & GUI-Actor-7B$^*$ + Verifier & - & -
    &
    & - & -
    &
    & - & -
    &
    & - & -
    &
    & - & -
    &
    & - & -
    &
    & - & - &  47.7 \\
    \midrule
    \multirow{11}{*}{\rotatebox{90}{\bf RL}}
    & GUI-G$^2$-3B$^*$~\cite{tang2025gui} & 20.3 & 9.4
    &
    & 42.2 & 2.1
    &
    & 43.4 & 2.8
    &
    & 43.8 & 10.0
    &
    & 46.3 & 15.1
    &
    & 29.0 & 4.5
    &
    & 37.6 & 6.0 &  25.5 \\
    & UI-R1-E-3B$^*$~\cite{lu2025ui} & 37.1 & 12.5
    &
    & 46.1 & 6.9
    &
    & 41.9 & 4.2
    &
    & 56.9 & 21.8
    &
    & 65.0 & 26.4
    &
    & 32.7 & 10.1
    &
    & - & - &  33.5 \\
    & InfiGUI-R1-3B$^*$~\cite{liu2025infigui} & 33.0 & 14.1
    &
    & 51.3 & 12.4
    &
    & 44.9 & 7.0
    &
    & 58.3 & 20.0
    &
    & 65.5 & 28.3
    &
    & 43.9 & 12.4
    &
    & 49.1 & 14.1 & 35.7 \\
    & SE-GUI-3B$^*$~\cite{yuan2025enhancing} & 38.1 & 12.5
    &
    & 55.8 & 7.6
    &
    & 47.0 & 4.9
    &
    & 61.8 & 16.4
    &
    & 59.9 & 24.5
    &
    & 40.2 & 12.4
    &
    & 50.4 & 11.8 & 35.9 \\
    & GUI-G1-3B$^*$~\cite{zhou2025gui} & 39.6 & 9.4
    &
    & 50.7 & 10.3
    &
    & 36.6 & 11.9
    &
    & 61.8 & 30.0
    &
    & 67.2 & 32.1
    &
    & 23.5 & 10.6
    &
    & 49.5 & 16.8 & 37.1 \\
    & InfiGUI-G1-3B$^*$~\cite{liu2025infigui} & 50.8 & 25.0
    &
    & 64.9 & 20.0
    &
    & 51.5 & 16.8
    &
    & 68.8 & 32.7
    &
    & 70.6 & 32.1
    &
    & 49.5 & 15.7
    &
    & - & - &  45.2 \\
    & UI-TARS-1.5-7B~\cite{qin2025ui}$^*$ & - & -
    &
    & - & -
    &
    & - & -
    &
    & - & -
    &
    & - & -
    &
    & - & -
    &
    & - & - &  49.6 \\
    & GroundNext-3B (RL)$^*$~\cite{feizi2025grounding} & 55.3&  32.8& &  65.6&  36.6&&   50.0&  24.5& &  66.0&  37.3 & & 74.6 & 50.9 & & 45.8 & 29.2& &  59.9&  33.6&  49.8\\
    & GTA1-7B$^*$~\cite{yang2025gta1guitesttimescaling} & \bf 66.9 & 20.7
    &
    & 62.6 & 18.2
    &
    & 53.3 & 17.2
    &
    & 76.4 & 31.8
    &
    & \bf82.5 & 50.9
    &
    & 48.6 & 25.9
    &
    & 65.5 & 25.2 & 50.1 \\
    &InfiGUI-G1-7B$^*$~\cite{liu2025infigui}&57.4& 23.4&& 74.7& 24.1&& 64.6& 15.4 &&\bf80.6& 31.8&& 75.7& 39.6 &&57.0& 29.2&& 68.4 &25.2&  51.9\\
    &GroundNext-7B (RL)$^*$~\cite{feizi2025grounding}&  50.2 & 34.3& &  73.4&  40.0& &  59.6&  23.8& &  70.1 & 42.7& &  74.6 & 54.7 & & 53.3 & 30.3 & & 60.5&  33.6&  52.9\\
    \midrule
    & \textbf{\method{}-3B-lite}$^*$ & 44.7 & 21.9
    &
    & 68.2 & 28.3
    &
    & 59.6 & 21.7
    &
    & 69.4 & 37.3
    &
    & 70.6 & 49.1
    &
    & 65.4 &  31.5
    &
    & 62.0 & 30.0 & 49.8 \\
    & \textbf{\method{}-3B-lite}$^*$+ \faSearchPlus & 54.3&  26.6& & \bf 77.9 & 37.2& &  67.2&  \underline{31.5}& &  76.4 & 39.1& &  \underline{80.8}&  \underline{64.2} & & 61.7 & 32.6& &  69.5 & 36.8 &  \underline{57.0}\\
    & \textbf{\method{}-3B}$^*$ & 50.3 & \underline{34.4}
    &
    & 68.8 & \underline{46.9}
    &
    & 52.0 & 30.1
    &
    & 75.0 & \underline{43.6}
    &
    & 74.0 & 50.9
    &
    & 56.1 & \underline{39.3}
    &
    & 62.1 & \underline{40.2} &  \bf53.8 \\
    & \textbf{\method{}-3B}$^*$+ \faSearchPlus & \underline{58.4} & \bf45.3
    &
    & 73.4 & \bf48.3
    &
    & 63.6 & \bf41.3
    &
    & 77.8 & \bf47.3
    &
    & \bf82.5 & \bf66.0
    &
    & 67.3 & \bf49.4
    &
    & 70.0 & \bf47.9 &  \bf61.5 \\
    \bottomrule
    \end{tabular}
    }
    \label{tab:screenspot_pro}
\end{table*}

\noindent\textbf{Baselines.}Three categories of methods are compared: 
(1) General MLLMs; (2) GUI-specific supervised fine-tuned models; (3) GUI-specific Reinforcement fine-tuned models. For our variants, \method{}-3B-lite is trained on the subset of GUI-Actor~\cite{wu2025gui}'s training data, while \method{}-3B additionally incorporates the 250k subset of GroundNext~\cite{feizi2025grounding}'s training data (710k). UI-Vision results are in-domain for both \method{}-3B and GroundNext.
\begin{table*}[!t]
    \centering
    \footnotesize
    \caption{Performance comparison on \textbf{ScreenSpot-v2} and \textbf{UI-Vision/MMBench-GUI-L2}. ``-'' indicates unreported results in original papers. Methods with $^*$ are Qwen2.5-VL-based.}
    \begin{subtable}[t]{0.48\textwidth}
        \centering
        \resizebox{\linewidth}{!}{
        \begin{tabular}{clccccccc}
    \toprule
    & \multirow{3}{*}{\textbf{Model}} 
      & \multicolumn{2}{c}{\textbf{Mobile}}
      & \multicolumn{2}{c}{\textbf{Desktop}}
      & \multicolumn{2}{c}{\textbf{Web}}
      & \multirow{2.5}{*}{\textbf{Avg.}} \\
    \cmidrule(lr){3-4}\cmidrule(lr){5-6}\cmidrule(lr){7-8}
    & & Text & Icon & Text & Icon & Text & Icon &  \\
    \midrule
    \multirow{6}{*}{\rotatebox{90}{\bf General}}
    & Operator~\cite{cua2025}  & 47.3 & 41.5 & 90.2 & 80.3 & 92.8 & 84.3 & 70.5\\
    & GPT-4o~\cite{openai2024gpt4o} + OmniParser-v2~\cite{yu2025omniparserv2structuredpointsofthoughtunified} & 95.5 & 74.6 & 92.3 & 60.9 & 88.0 & 59.6 & 80.7 \\
    & Qwen2.5-VL-3B~\cite{bai2025qwen2} & 93.4 & 73.5 & 88.1 & 58.6 & 88.0 & 71.4 & 80.9 \\
    & Qwen2.5-VL-7B~\cite{bai2025qwen2} &97.6 & 87.2 & 90.2 & 74.2 & 93.2 & 81.3 & 88.8   \\
    &Qwen3-VL-8B~\cite{bai2025qwen3} &\bf 99.7 &87.7 &94.8 &83.6& \underline{95.3}& 85.7& \underline{92.1}\\
    &Qwen3-VL-30B-A3B~\cite{bai2025qwen3}& 99.0& 87.7& 95.4& 82.9& \underline{95.3} &83.7 &91.7\\
    \midrule
    \multirow{7}{*}{\rotatebox{90}{\bf SFT}}
    & OS-Atlas-7B~\cite{wu2024atlas} & 95.2 & 75.8 & 90.7 & 63.6 & 90.6 & 77.3 & 84.1\\
    & UGround-V1-7B~\cite{gou2024navigating} & 95.0 & 83.3 & 95.0 & 77.8 & 92.1 & 77.2 & 87.6 \\
    & UI-TARS-7B~\cite{qin2025ui} & 96.9 & \underline{89.1} & 95.4 & 85.0 & 93.6 & 85.2 & 91.6\\
    & \textsc{Jedi}-3B$^*$~\cite{xie2025scaling}  & 96.6 & 81.5 & \bf96.9 & 78.6 & 88.5 & 83.7 & 88.6\\
    & \textsc{Jedi}-7B$^*$~\cite{xie2025scaling} & \bf96.9 & 87.2 & 95.9 & 87.9 & 94.4 & 84.2 & 91.7 \\
    & GUI-Actor-3B~\cite{wu2025gui}& 97.6& 83.4& \bf96.9& 83.6& 94.0& 85.7& 91.0 \\
    &GUI-Actor-3B + Verifier& 98.3 & 85.3&\bf96.9 & 87.9 & 95.3 & \bf86.7 & \bf92.4\\
    \midrule
    \multirow{7}{*}{\rotatebox{90}{\bf RL}}
    & GUI-G$^2$-3B$^*$~\cite{tang2025gui} &96.5 &82.5&95.4&75.0&88.4&72.4&86.3\\
    &InfiGUI-R1-3B$^*$~\cite{liu2025infigui} &97.1 &81.2 &94.3& 77.1& 91.7& 77.6 &87.5\\
    &GroundNext-3B (RL)$^*$~\cite{feizi2025grounding} &94.8& \bf96.4 &93.9 &87.1 &90.6 &79.3& 88.5\\
    & UI-R1-E-3B$^*$~\cite{lu2025ui}  &98.2 &83.9& 93.2 &83.7& 94.8 &75.0 &89.5\\
    & UI-TARS-1.5-7B$^*$~\cite{qin2025ui} &95.9 &84.8 &94.9& 80.7& 90.6& \underline{86.2}& 89.7\\
    &GroundNext-7B (RL)$^*$~\cite{feizi2025grounding}& 96.6 &88.2& 95.4 &87.9& 94.9& 75.9 &90.4\\
    &InfiGUI-G1-3B$^*$~\cite{liu2025infigui} & 99.3 &88.2 &94.8& 82.9 &94.9& 80.3& 91.1\\ 
    \midrule
    & \textbf{\method{}-3B-lite}$^*$ &99.2&85.9&\underline{96.1}&\underline{88.9}&\bf96.1&80.2&91.5\\
    & \textbf{\method{}-3B}$^*$ &\underline{99.6}&84.8&95.6&\bf91.3&94.3&84.8&\underline{92.1}\\
    \bottomrule
    \end{tabular}}
    \caption{ScreenSpot-v2.}
    \label{tab:screenspot_v2}
    \end{subtable}
    \hfill
    \begin{subtable}[t]{0.51\textwidth}
        \centering
        \resizebox{\linewidth}{!}{
        \begin{tabular}{clcccc|ccc}
    \toprule
    & \multirow{2}{*}{\textbf{Model}} 
      & \multicolumn{4}{c|}{\textbf{UI-Vision}}& \multicolumn{3}{c}{\textbf{MMBench-GUI-L2}}\\
      \cmidrule(lr){3-6}\cmidrule(lr){7-9}
    & & Basic & Func. & Spatial &\bf Avg. &Basic &Advanced&\bf Avg. \\
    \midrule
    \multirow{4}{*}{\rotatebox{90}{\bf General}}
    &Qwen2.5-VL-7B$^*$~\cite{bai2025qwen2} &1.2& 0.8& 0.5& 0.9&38.0   & 29.8    & 33.9\\
    & Qwen3-VL-2B$^*$~\cite{bai2025qwen3} & 16.4 & 19.1 & 4.6  & 13.1 & 84.3  &54.9 &69.5 \\
    & Qwen3-VL-8B$^*$~\cite{bai2025qwen3} & 27.8 & 29.6 & 9.4  & 21.9 &90.2&72.7&81.4 \\
    & Qwen3-VL-30B-A3B$^*$~\cite{bai2025qwen3} & 31.2 & 31.9 & 14.6 & 25.6&91.0&76.5&83.7 \\
    \midrule
    \multirow{7}{*}{\rotatebox{90}{\bf SFT}}
    &OS-Atlas-Base-7B~\cite{wu2024atlas}    &12.2& 11.2& 3.7& 9.0  &       52.8   &   30.1 &      41.4\\
     &Aguvis-7B~\cite{xu2024aguvis} &17.8& 18.3& 5.1 &13.7&-  &   - &      45.7 \\
    &UGround-V1-7B~\cite{gou2024navigating} &15.4 &17.1 &6.3 &12.9  &               78.4  &  53.0&     65.7 \\
    & GUI-Actor-3B$^*$~\cite{wu2025gui} &-&-&-& 19.7&-&-& 69.8\\
    & GUI-Actor-7B$^*$~\cite{wu2025gui} &-&-&-&21.9 &-&-&70.9\\
    &JEDI-7B$^*$~\cite{xie2025scaling}&-&-&-&24.8 &-&-&70.4\\
    &GroundNext-3B (SFT)$^*$~\cite{feizi2025grounding}&70.9& 59.8 &45.1 &58.2 &86.5&64.5&75.5\\
    \midrule
     \multirow{8}{*}{\rotatebox{90}{\bf RL}}
    &UI-TARS-1.5-7B$^*$~\cite{qin2025ui}&- &-  &  -   &  20.8 & 78.4   & 50.4   &  64.3 \\
    &InfiGUI-G1-3B$^*$~\cite{liu2025infigui}&31.2& 28.0& 8.2& 22.0 &84.1 &  62.9  &  73.4    \\
    &InfiGUI-G1-7B$^*$~\cite{liu2025infigui}&36.2& 31.9& 11.5& 26.1& 88.3&   73.3&    80.8\\
    &GTA1-7B$^*$~\cite{yang2025gta1guitesttimescaling}&- &-  &  -   &  25.7 &-  &  -    & 79.4 \\
    &GUI-G$^2$-7B$^*$~\cite{tang2025gui} &- &-  &  -   &  25.6 &-  &  -    & 79.5 \\
    &Holo2-8B~\cite{hcompany2025holo2} & 43.6 & 43.5 & 19.7 & 35.1 & \underline{91.7}   &\underline{77.4} &   \underline{84.5}\\
    &Holo2-30B-A3B~\cite{hcompany2025holo2}& 51.0 & 50.1 & 23.2 & 40.9&\bf92.8&\bf\bf80.7 &\bf86.8 \\
    &GroundNext-3B (RL)$^*$~\cite{feizi2025grounding}&\bf72.9 &\bf63.9 &\bf50.6& \bf62.1 &87.3&67.1&77.1\\
    \midrule
    & \textbf{\method{}-3B-lite}$^*$ &39.2 & 33.9 & 10.6 & 27.9 &86.2  &66.5 & 76.3 \\
    & \textbf{\method{}-3B}$^*$ & \underline{70.8}& \underline{63.1}& \underline{46.2} & \underline{60.0}&88.3  &69.9 &  79.1  \\    
    \bottomrule
    \end{tabular}}
    \caption{UI-Vision and MMBench-GUI-L2.}
    \label{tab:ui_mm}
    \end{subtable}
    \label{tab:ui_mm_v2}
\end{table*}
\subsection{Main Results: Grounding Performance}
We compare \method{} with state-of-the-art methods on ScreenSpot-Pro~(\linebreak\cref{tab:screenspot_pro}), ScreenSpot-v2~(\cref{tab:screenspot_v2}), UI-Vision, MMBench-GUI-L2~(\cref{tab:ui_mm}), and OSWorld-G~(\cref{tab:osworld_g}).

\textbf{Desktop scenario.}
\method{} achieves strong performance on challenging high-resolution desktop benchmarks. On ScreenSpot-Pro, \method{}-3B-lite reaches 49.8\% accuracy, and additional desktop-focused training with 200k GroundCUA instructions further scales \method{}-3B to 53.8\%, outperforming strong SFT and RL baselines. Moreover, our patch-level attention grounding naturally enables a two-step zoom-in inference, which substantially improves performance to 57.0\% for \method{}-3B-lite and 61.5\% for \method{}-3B, with particularly large gains on the abstract icon subset. Despite being trained with less than one-third of the GroundCUA data used by GroundNext-3B, \method{}-3B achieves comparable in-domain performance on UI-Vision, while outperforming it on the other benchmarks. On OSWorld-G,  given already a competitive base model (62.8\%), \method{}-3B is further improved to 68.1\% with zoom-in, performing competitively against the strongest large-scale baselines (GTA1-7B and Qwen3-VL-30B-A3B). These results highlight the practical advantage of directly supervising intrinsic attention for dense, high-resolution desktop interfaces.

\textbf{General grounding across domains.}
On ScreenSpot-v2, \method{}-3B outperforms same-scale and larger baselines, and remains competitive with the two-step GUI-Actor-3B with verifier, which is trained on 9.6M samples, demonstrating the data efficiency of \method{}. Notably, \method{} achieves strong icon grounding across desktop domains, indicating robust generalization beyond text-heavy cases. On MMBench-GUI-L2, \method{}-3B consistently outperforms same-scale baselines and matches many larger models.

Overall, \method{} is both effective and scalable. The intrinsic attention grounding manner provides data efficiency for better scaling and flexibility for inference-time improvements without additional training. The consistent improvement from \method{}-3B-lite to \method{}-3B with additional single-domain data further demonstrates the scalability of intrinsic attention supervision. The improvement of \method{} over GUI-Actor shows the advantage of intrinsic attention supervision without adding an extra module and training stage, with much less training effort in a natural manner.
 \begin{table*}[!t]
    \centering
    \footnotesize
    \caption{Performance comparison of different models across various task categories based on Mobile, Desktop, Web and Average scores on \textbf{OSWorld-G-Refine}. Methods with $^*$ are Qwen2.5-VL-based. \faSearchPlus denotes two-step inference with zoom-in.}
    \resizebox{0.8\linewidth}{!}{
    \begin{tabular}{clccccc} 
    \toprule
    & \textbf{Model}
      & \makecell{\textbf{Text}\\\textbf{Matching}} 
      & \makecell{\textbf{Element}\\\textbf{Recognition}} 
      & \makecell{\textbf{Layout}\\\textbf{Understanding}}
      & \makecell{\textbf{Fine-grained}\\\textbf{Manipulation}} 
      & {\textbf{Avg.}} \\
    \midrule 
    \multirow{6}{*}{\rotatebox{90}{\bf General}}
    & Operator~\cite{cua2025}  &  51.3& 42.4& 46.6& 31.5& 40.6  \\
    & Gemini-2.5-Pro~\cite{comanici2025gemini} & 59.8 &45.5& 49.0& 33.6 &45.2 \\
    & Qwen2.5-VL-3B~\cite{bai2025qwen2} &41.4 &28.8 &34.8 &13.4& 27.3\\
    & Qwen2.5-VL-7B~\cite{bai2025qwen2} & 45.6& 32.7 &41.9 &18.1 & 31.4  \\
    &Qwen3-VL-8B~\cite{bai2025qwen3} &78.2& 71.5 &72.3& 55.7&67.0\\
    &Qwen3-VL-30B-A3B~\cite{bai2025qwen3} &77.8 &\underline{75.8} &\underline{74.7}& 54.4& \bf69.3\\
    \midrule
    \multirow{7}{*}{\rotatebox{90}{\bf SFT}}
    & OS-Atlas-7B~\cite{wu2024atlas} & 44.1 &29.4& 35.2 &16.8 & 27.7 \\
    & UGround-V1-7B~\cite{gou2024navigating} & 51.3 &40.3& 43.5& 24.8 &36.4  \\
    & UI-TARS-7B~\cite{qin2025ui} & 60.2& 51.8& 54.9 &35.6 &47.5 \\
    & \textsc{Jedi}-3B$^*$~\cite{xie2025scaling} & 67.4 &53.0 &53.8 &44.3 &50.9 \\
     & \textsc{Jedi}-7B$^*$~\cite{xie2025scaling} &65.9& 55.5& 57.7 &46.9 & 54.1  \\
     & GUI-Actor-3B$^*$~\cite{wu2025gui}  &64.4&   60.6&  64.8  &33.6   &54.6	\\
     & GUI-Actor-7B$^*$~\cite{wu2025gui}  &65.9&	62.7&	66.4&	38.2	&56.6	\\
    \midrule
    \multirow{6}{*}{\rotatebox{90}{\bf RL}}
    &SE-GUI-7B$^*$~\cite{yuan2025enhancing} &-&-&-&-&33.9\\
    &InfiGUI-G1-3B$^*$~\cite{liu2025infigui}& 65.5& 53.0& 56.1& 34.2&49.6\\
    &InfiGUI-G1-7B$^*$~\cite{liu2025infigui}& \underline{72.0}& 63.6& 66.8& 46.3 &59.9\\
    &GUI-G$^2$-7B$^*$~\cite{tang2025gui} &-&-&-&-&61.9\\
    &UI-TARS-1.5-7B$^*$~\cite{qin2025ui} & 52.6& 75.4& 72.4& \underline{66.7}& 64.2 \\
    &GTA-1-7B$^*$~\cite{yang2025gta1guitesttimescaling}& 63.2& \bf82.1 &74.2 &\bf70.5 &  67.7\\
    \midrule
    & \textbf{\method{}-3B-lite}$^*$ &64.8&65.5&68.8&36.8&58.3\\
    & \textbf{\method{}-3B-lite}$^*$+ \faSearchPlus&71.3&70.6&73.1&47.4&63.8 \\
    & \textbf{\method{}-3B}$^*$&67.4&	71.2&	72.7&	39.5&	62.8 \\
    & \textbf{\method{}-3B}$^*$+ \faSearchPlus&\bf72.4&	\underline{75.8}&	\bf78.7&	48.7&	\underline{68.1} \\
    \bottomrule
    \end{tabular}}
    \label{tab:osworld_g}
\end{table*}

\subsection{Ablations}\label{sec:ablation}
\cref{tab:ablation} shows the ablation results on ScreenSpot-v2 and ScreenSpot-Pro. All the ablation variants are trained on the 45k ablation training data with Qwen2.5-VL-3B-Instruct as the backbone. GUI-Actor (45k) is trained in two stages same as the original setting.

\noindent\textbf{Compare Coordinate-free modeling manners.} In \cref{tab:ablation}, we include 3 different coordinate-free GUI grounding manners, one embedding-based: GUI-Actor, which relies on hidden embeddings for computing similarity with extra modules, and two attention-based: vanilla attention grounding as \cref{eq:aggre_vanilla} and simplified attention grounding as \cref{eq:aggre_final}. With the same importance on all the query tokens and attention heads, the vanilla attention grounding can converge faster than GUI-Actor on more complex visual grounding tasks in ScreenSpot-Pro. When simplifying the aggregation on query tokens, even with naive "weighting uniformly", the simplified attention grounding can have \textbf{1.50\%} and \textbf{4.24\%} improvement over GUI-Actor for ScreenSpot-v2 and ScreenSpot-Pro. These results demonstrate the effectiveness and faster convergence of attention-based visual grounding methods than embedding-based methods, and show the advantage of compressing query contexts into $\texttt{<ANCHOR>}$ token instead of manually controlling the grounding importance from each query token. 
\begin{table*}[!t]
    \centering
    \footnotesize
    \begin{minipage}[c]{0.56\textwidth}
        \centering
    \caption{Ablation results on ScreenSpot-v2 and ScreenSpot-Pro of coordinate-free GUI grounding methods fine-tuned with a 45k ablation dataset. Variants in \colorbox{blue!10}{blue} represent the selected settings for \method{}.}
    \resizebox{\linewidth}{!}{
    \begin{tabular}{p{160pt}cccccc}
        \toprule
        \multirow{2}{*}{\textbf{Model}} & \multicolumn{3}{c@{\hspace{1em}}}{\textbf{ScreenSpot-v2}} & \multicolumn{3}{c@{\hspace{1em}}}{\textbf{ScreenSpot-Pro}}\\
        \cmidrule(lr){2-4}\cmidrule(lr){5-7}
        & Text & Icon & \textbf{Avg.}& Text & Icon & \textbf{Avg.} \\
        \midrule
        \rowcolor{gray!30} 
        \multicolumn{7}{l}{\textit{Existing Coordinate-free GUI Grounding}} \\
        GUI-Actor-3B (45k) & 96.24 & 75.99 & 87.42 & 50.67 & 12.25 & 35.99 \\
        Vanilla Attention Grounding (45k) & 95.68 & 75.45 & 86.87 & 53.02 & 12.42 & 37.51 \\
        \midrule
        \rowcolor{gray!30} 
        \multicolumn{7}{l}{\textit{Simplified Attention Grounding: GUI-AIMA-3B (45k)}} \\
        \rowcolor{gray!10}\multicolumn{7}{l}{\textit{Attention Head Weighting without $\mathcal{Q}_s$} in \cref{eq:unified_head_score}}\\
        + weighting uniformly & 97.08 & 78.34 & 88.92 & 56.50 & 13.91 & 40.23 \\
        + weighting with all $\mathcal{Q}'=\mathcal{Q}$ & 96.24 & 78.34 & 88.44 & 52.61 & 13.41 & 37.63 \\
        + weighting with $\mathcal{Q}'=\{\texttt{<ANCHOR>}\}$ & 96.66 & 76.35 & 87.81 & 56.91 & 14.57 & 40.73 \\
        \rowcolor{gray!10}\multicolumn{7}{l}{\textit{Attention Head Weighting with }$\mathcal{Q}'=\mathcal{Q}_s$\textit{ in} \cref{eq:unified_head_score}}\\
        + layer-wise $\operatorname{top-1} \mathcal{Q}_s^l$ & 96.52 & \textbf{81.23} & \textbf{89.86} & 57.32 & \underline{15.73} & 41.43 \\
        + layer-wise $\operatorname{top-3} \mathcal{Q}_s^l$ & 96.66 & 79.06 & 88.99 & 57.83 & \underline{15.07} & 41.49 \\
        \rowcolor{blue!10}
        + global $\operatorname{top-1} \mathcal{Q}_s$ & \textbf{97.49} & 78.88 & 89.39 & \underline{59.26} & 14.40 & \underline{42.13} \\
        + global $\operatorname{top-3} \mathcal{Q}_s$ & 95.82 & 70.94 & 84.98 & 57.01 & 14.40 & 40.73 \\
        \rowcolor{blue!10}
        ~~~~ + weighted patch-wise labeling & \underline{97.21} & \underline{79.60} & \underline{89.54} & \textbf{61.00} & 14.90 & \textbf{43.39} \\
        \bottomrule
      \end{tabular}}
    \label{tab:ablation}
\end{minipage}
    \hfill
    \begin{minipage}[c]{0.42\textwidth}
    \begingroup
    \makeatletter\renewcommand{\@captype}{figure}\makeatother
    \includegraphics[width=\linewidth]{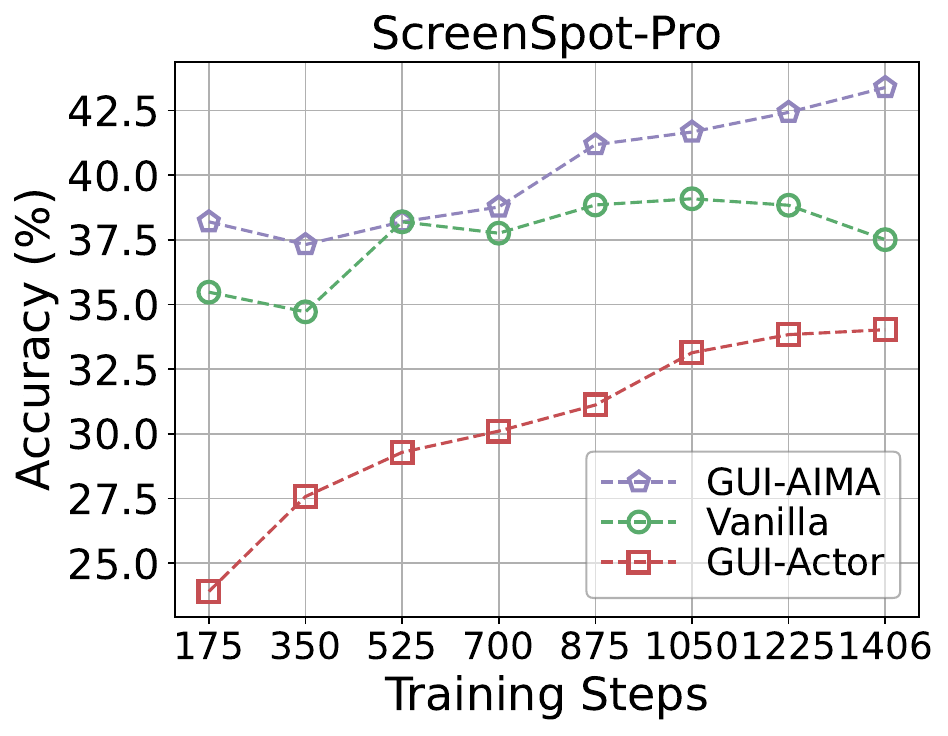}
    \caption{Performance curves over training steps on ScreenSpot-Pro for \method{}, GUI-Actor, and vanilla attention grounding trained on the 45k ablation dataset.}
    \label{fig:Convergence}
    \endgroup
    \end{minipage}
\end{table*}
\begin{table*}[!t]
\centering
\caption{Inference overhead on OSWorld-G. AR ($\texttt{<ANCHOR>}$) is constrained to generate the same $\texttt{<ANCHOR>}$ sequence as \method{}, isolating the grounding attention recomputation. AR (coordinate) denotes coordinate-based grounding with a longer sequence.}
\label{tab:inference_overhead}
\setlength{\tabcolsep}{3.0pt}
\resizebox{1\linewidth}{!}{
\begin{tabular}{lcccccccc}
\toprule
& \multicolumn{2}{c}{Ours (FlashAttention + partial Eager)} & \multicolumn{2}{c}{Ours (full Eager)} & \multicolumn{2}{c}{AR ($\texttt{<ANCHOR>}$)} & \multicolumn{2}{c}{AR (coordinate)} \\
\cmidrule(lr){2-3}\cmidrule(lr){4-5}\cmidrule(lr){6-7}\cmidrule(lr){8-9}
Setting & Time (ms) & Max Mem. (GB)&Time (ms) & Max Mem. (GB)&Time (ms) & Max Mem. (GB)&Time (ms) & Max Mem. (GB) \\
\midrule
Single-step & \underline{778.5} & \underline{10.69} & 4548.1 & 24.75 & \bf{723.6} & \bf{8.40} & 1135.3 & \bf{8.39} \\
Two-step & \underline{1016.5} & \underline{10.69} & 5032.7 & 24.75 & -- & -- & -- & -- \\
\bottomrule
\end{tabular}
}
\end{table*}

\noindent\textbf{GUI grounding without Visual-sink Query Token $\mathcal{Q}_s$.} For attention-based GUI grounding variants, weighting the query-visual correlation of each head via the cumulative sum of all query's predictions as $\mathcal{Q}'=\mathcal{Q}$ (TAG's manner) or only the prediction from $\texttt{<ANCHOR>}$ as $\mathcal{Q}'=\{\texttt{<ANCHOR>}\}$ in \cref{eq:unified_head_score} leads to more biased attention predictions than uniform weighting, with worse ScreenSpot-v2 performance of both variants and no improvement on ScreenSpot-Pro as shown in ``Attention Head Weighting without $\mathcal{Q}_s$ in \cref{eq:unified_head_score}'' part of \cref{tab:ablation}. This bias comes from the misleading weighting from visual-unrelated tokens for the variants using all $\mathcal{Q}$ and from using the premature $\texttt{<ANCHOR>}$ token for grounding in the later variant. 

\noindent\textbf{The effect of Visual-sink Query Token $\mathcal{Q}_s$.} For ablation of using $\mathcal{Q}_s$ for $w_{l,h}$, we vary the $\operatorname{K}$ value between 1 and 3 in the $\operatorname{topK}$ selection of $\mathcal{Q}_s$ tokens and explore whether differing $\mathcal{Q}_s^l$ for each MLLM's layer, i.e., $\mathcal{Q}_s^l\coloneqq \operatorname*{arg\,topK}_{q_i\in\mathcal{Q}}(c_{q_i}^{\,l})$, or using the same $\mathcal{Q}_s$ for all layers as in \cref{eq:find_sink}. In \cref{tab:ablation}, we can observe that the layer-wise manner performs slightly better on the icon-based tasks, but uniformly-selected same $\mathcal{Q}_s$ for all layers leads to more balanced results on both text and icon sets. Among both manners, $\operatorname{top-1}$ selection remains the overall best performance. These ablations verify the ``global $\operatorname{top-1} \mathcal{Q}_s$'' setting of \method{}, with \textbf{1.90\%} improvement over ``weighting uniformly'' on ScreenSpot-Pro.
\begin{table*}[!t]
    \centering
    \footnotesize
    \begin{minipage}[c]{0.42\textwidth}
    \begin{minipage}[t]{\textwidth}
    \centering
    \caption{Sensitivity to Gaussian scaling factor $\alpha$ in \cref{eq:labeling}.}
    \label{tab:alpha_sensitivity}
    \resizebox{\linewidth}{!}{
    \begin{tabular}{lccccc}
    \toprule
     $\alpha$ & 0.6 & 0.7 & 0.8 & 0.9 & 1.0 \\
    \midrule
    ScreenSpot-v2 & 87.34 & 89.15 & \textbf{89.54} & 89.47 & 88.60 \\
    ScreenSpot-Pro & 39.91 & 41.62 & \textbf{43.39} & 42.88 & 42.25 \\
    \bottomrule
    \end{tabular}
    }
    \end{minipage}
    ~\\
    \begin{minipage}[t]{\textwidth}
    \caption{Cross-model validation on InternVL3.5 trained with 45K data.}
    \label{tab:cross_model}
    \resizebox{\linewidth}{!}{
    \begin{tabular}{lcccc}
    \toprule
    Model & SSv2 & SSPro & OSW-G & Overall \\
    \midrule
    InternVL3.5-2B & 79.6& 12.2 & 27.7 & 39.8 \\
    \method{}-InternVL3.5-2B & \underline{86.7}& 16.0 &\underline{33.9} &45.5 \\
    InternVL3.5-4B & 85.1 & \underline{18.1} & \underline{33.9} & \underline{45.7} \\
    \method{}-InternVL3.5-4B & \textbf{88.8} & \textbf{19.9} & \textbf{36.4} & \textbf{48.4}\\
    \bottomrule
    \end{tabular}
    }
    \end{minipage}
    \end{minipage}
    \hfill
    \begin{minipage}[c]{0.56\textwidth}
        \centering
    \caption{Analysis experiment results on ScreenSpot-pro. \textbf{Relax@k} measures how many previously incorrect offsets are recovered when the ground-truth bounding box is expanded by k visual patches along each dimension. \textbf{Corrected} refers to the number of offsets corrected by the second step, while \textbf{Lost} refers to the number of predictions that degrade after the second step.}
    \resizebox{\linewidth}{!}{
    \begin{tabular}{lcccc|cc|c}
        \toprule
        Model& Relax@1 & Relax@3 & Relax@5 & \textbf{Total} & \textbf{Corrected} & \textbf{Lost}&\textbf{Improved} \\
        \midrule
        \rowcolor{gray!30}
        \multicolumn{8}{l}{\textit{1-step}} \\
        Origin 1-step & 158 & 92 & 56 & 306 & - & - &-\\
        \midrule
        \rowcolor{gray!30}
        \multicolumn{8}{l}{\textit{2-step}} \\
        Crop-only & 123 & 83 & 43 & 249 & 156 & 107 &49\\
        Crop + 1.5$\times$Zoom-in & 47 & 74 & 38 & 159 & 208 & 48 & 160\\
        Crop + 2$\times$Zoom-in& 39 & 63 & 45 & 147 & 215&33& \bf 182  \\
        Crop + 3$\times$Zoom-in& 39 & 69 & 44 & 152 &219 & 42&177 \\
        Crop + 4$\times$Zoom-in& 31 & 72 & 38 & 141 &224 & 48&176\\
        \bottomrule
      \end{tabular}}
    \label{tab:analysis_zoom}
\end{minipage}
\end{table*}

\noindent\textbf{Patch-wise labeling considering overlapping and distance.} Here, we also justify the overlapping- and distance-aware~\cite{tang2025gui} patch-wise labeling for fine-tuning \method{} in \cref{eq:labeling}. Based on ``global $\operatorname{top-1} \mathcal{Q}_s$'', down-weighting partial positive and distanced image patches instead of labeling all patches equally leads to \textbf{1.26\%} further enhancement on ScreenSpot-Pro.
We further report the sensitivity to the Gaussian scaling factor $\alpha$ in \cref{eq:labeling} in \cref{tab:alpha_sensitivity}, supporting our selection of $\alpha=0.8$.

\noindent\textbf{Cross-model validation.} To validate the generality of \method{} beyond Qwen2.5-VL, we apply it to InternVL3.5~\cite{wang2025internvl3}'s 2B and 4B models trained with the same 45K ablation data, where the main adaptation is to map visual patch indices back to spatial coordinates under InternVL3.5's dynamic image tiling. As shown in \cref{tab:cross_model}, \method{} consistently improves over the corresponding backbones, and the adapted 2B model is competitive with the 4B backbone.

\noindent\textbf{Training and inference efficiency.} In \cref{fig:Convergence} , we show the convergence tendency of three methods: \method{}, vanilla attention grounding and GUI-Actor (with extra warm-up stage already) initialized from Qwen-2.5-VL-3B. \method{} converges fastest and achieves the best final results, with steady improvements. Vanilla attention grounding fluctuates as supervision on all tokens from the pretrained vocabulary impairs the general capacity. And GUI-Actor needs a longer training period to customize its extra modules for GUI grounding.
For inference overhead, we report detailed comparisons in \cref{tab:inference_overhead} on OSWorld-G using a NVIDIA A6000 GPU. For a fair comparison, AR ($\texttt{<ANCHOR>}$) is constrained to generate the same $\texttt{<ANCHOR>}$ sequence as \method{}, isolating the grounding attention recomputation. AR (coordinate) denotes coordinate-based grounding with a longer coordinate sequence. Our implementation is time-efficient but more memory-consuming than AR (coordinate), while much more efficient than full eager attention.
\subsection{Analysis of Two-step Inference with Zoom-in}\label{sec:zoomin_analysis}
In \cref{tab:analysis_zoom}, we present the analysis results of the two-step inference with zoom-in introduced in \cref{sec:zoom}. The analysis focuses on incorrect offset predictions, where the predicted center points fall outside the ground-truth bounding box by a small degree. \cref{tab:analysis_zoom} reports the statistics of offset errors for one-step and two-step inference with different offset distances, and the number of {Corrected} predictions as well as the {Lost} originally-correct predictions after applying the second crop-and-zoom-in step. 
We observe that performing inference on the focused region determined by the first-step prediction already corrects more than a 18.6\% samples (57) with ``Crop-only'' operation, arising from fewer irrelevant patches showing as distracting noise for our patch-wise grounding. And after cropping, zooming in by a factor from 1.5× to 4× significantly reduces offset errors, especially those with slight deviation (Relax@1). It also helps resolving large-offset errors, achieving up to 31\% (29 fixed) and 32\% (18 fixed) error reduction for Relax@3 and Relax@5, respectively. From the analysis results, we find that the performance of two-stem inference is not very sensitive to the zoom-in factor, while a 2-time zoom-in provides the best performance.
\section{Conclusion}
We present \method{}, an attention-based, coordinate-free framework for GUI visual grounding that directly supervises intrinsic multi-head self-attention with patch-wise grounding signals. \method{} replaces impractical vanilla token-wise attention aggregation with an anchored-attention formulation and improves multi-head aggregation via query-aware head weighting guided by visual-sink query tokens. Together with a click-aligned patch-wise labeling and a plug-and-play two-step zoom-in inference, \method{} provides a lightweight and scalable approach to grounding without introducing extra grounding modules. Across ScreenSpot-Pro, ScreenSpot-v2, OSWorld-G, UI-Vision, and MMBench-GUI-L2, \method{} achieves strong performance at the 3B scale, demonstrating both data efficiency on a small training set and consistent gains when scaled with additional desktop-domain data. Ablations further validate the contribution of each component, including anchored aggregation, head weighting, and labeling design. Future work includes extending attention-supervised grounding to more general visual grounding tasks and complex interaction scenarios.
\section*{Acknowledgements}
This work is partially supported under the AI Research Institutes program by National Science Foundation and the Institute of Education Sciences, U.S. Department of Education, through Award \# 2229873 – National AI Institute for Exceptional Education, NSF RI-2223292, an Amazon research award, and an Adobe gift fund. Any opinions, findings and conclusions or recommendations expressed in this material are those of the author(s) and do not necessarily reflect the views of the National Science Foundation, the Institute of Education Sciences, or the U.S. Department of Education. We thank Tong Sun for the discussion.
\newpage
\bibliographystyle{splncs04}
\bibliography{main}

\newpage
\begin{center}
{\Large\bfseries Supplementary Material}
\end{center}
\appendix
In the supplementary material, we provide extra experimental results (\cref{app:extra_res}, including running time), ablations on a larger training set (\cref{app:extra_abla}), attention head contribution analysis (\cref{app:attn_contribution}), further analysis on $\mathcal{Q}_s$ (\cref{app:qs_analysis}), additional baseline (\cref{app:baseline}) and implementation (\cref{app:anchor}) details. We also provide visualization examples (\cref{app:example}) and additional failure cases (\cref{app:failure}).
\section{Extra Results} \label{app:extra_res}
In \cref{tab:ssv1},  we show the result of \method{} and baselines on ScreenSpot-v1~\cite{cheng2024seeclick}. In \cref{tab:time},  we compare the inference and training time between \method{}-3B and GUI-Actor-3B.

\section{Extra Ablations} \label{app:extra_abla}
In \cref{tab:results_with_verifier}, we compare the performance of \method{} and GUI-Actor, with both combined with the multi-region verifier proposed by GUI-Actor, showing that \method{} can benefit to the same extent as GUI-Actor from the extra verifier.  In~\cref{tab:ablation_all}, we add extra ablation results with variants trained on the 85k training set of \method{}-lite. It includes the results of uniform multi-head weighting with/without $\texttt{<ANCHOR>}$ token, vanilla attention grounding using multi-head weighting with $\mathcal{Q}'=\mathcal{Q}$ in \cref{eq:unified_head_score}, \method{}-3B-lite with stop-gradient on multi-head weights, GUI-Actor-3B with/without distance-aware labeling. The extra ablation results using a larger ablation training set are consistent with the results in \cref{tab:ablation}.

\begin{table*}[!t]
    \centering
    \footnotesize
    \caption{Performance comparison of different models across various task categories based on Mobile, Desktop, Web and Average scores on \textbf{ScreenSpot-v1}.}
    \resizebox{0.9\linewidth}{!}{
    {
    \begin{tabular}{clccccccc}
        \toprule
        & \multirow{3}{*}{\textbf{Model}} 
          & \multicolumn{2}{c}{\textbf{Mobile}}
          & \multicolumn{2}{c}{\textbf{Desktop}}
          & \multicolumn{2}{c}{\textbf{Web}}
          & \multirow{2.5}{*}{\textbf{Avg.}} \\
        \cmidrule(lr){3-4}\cmidrule(lr){5-6}\cmidrule(lr){7-8}
        & & Text & Icon & Text & Icon & Text & Icon &  \\
        \midrule
        &GPT-4 & 22.6 & 24.5 & 20.2 & 11.8 & 9.2 & 8.8 & 16.2 \\
        &GPT-4o & 20.2 & 24.9 & 21.1 & 23.6 & 12.2 & 7.8 & 18.3 \\
        &Claude Computer Use & - & - & - & - & - & - & 83.0 \\
        &Gemini 2.0 & - & - & - & - & - & - & 84.0 \\
        \midrule
        \multirow{6}{*}{\rotatebox{90}{\bf 3B}}
        &UGround-V1-2B & 89.4 & 72.0 & 88.7 & 65.7 & 81.3 & 68.9 & 77.7\\
        &UI-TARS-2B & 93.0 & 75.5 & 90.7 & 68.6 & 84.3 & 74.8 & 82.3 \\
        &GUI-Actor-2B (Qwen2-VL) & 93.0 & 79.9 & 88.1 & 78.6 & 90.9 & 84.0 & 86.5 \\
        &GUI-Actor-3B (Qwen2.5-VL) & 94.5&83.8&92.8&82.1&91.3&82.5&88.4 \\
        \cmidrule(lr){2-9}
        &\textbf{\method{}-3B-lite} & 97.1&82.1&95.4&83.6&90.9&80.6&\textbf{88.8} \\
        \midrule
        \multirow{13}{*}{\rotatebox{90}{\bf 7B}}
        &Qwen2-VL-7B & 75.5 & 60.7 & 76.3 & 54.3 & 35.2 & 25.7 & 55.3 \\
        &CogAgent-7B & 67.0 & 24.0 & 74.2 & 20.0 & 70.4 & 28.6 & 47.4 \\
        &SeeClick-9.6B & 78.0 & 52.0 & 72.2 & 30.0 & 55.7 & 32.5 & 53.4 \\
        &Magma-8B & 60.4 & 58.5 & 75.3 & 52.9 & 69.1 & 52.0 & 60.3\\
        &Aguvis-G-7B & 88.3 & 78.2 & 88.1 & 70.7 & 85.7 & 74.8 & 81.8 \\
        &OS-Atlas-7B & 93.0 & 72.9 & 91.8 & 62.9 & 90.9 & 74.3 & 82.5 \\
        &Aguvis-7B & 95.6 & 77.7 & 93.8 & 67.1 & 88.3 & 75.2 & 84.4 \\
        &UGround-v1-7B & 93.0 & 79.9 & 93.8 & 76.4 & 90.9 & 84.0 & 86.3 \\
        &SparkUI-Parser&94.9&83.8&95.9&80.7&89.6&82.9&88.0\\
        &UI-TARS-7B & 94.5 & 85.2 & 95.9 & 85.7 & 90.0 & 83.5 & 89.5 \\
        &GUI-Actor-7B (Qwen2-VL)& 94.9 & 82.1 & 91.8 & 80.0& 91.3 & 85.4 & 88.3 \\
        &GUI-Actor-7B (Qwen2.5-VL) & 96.3&85.2&95.4&82.9&90.4&85.4&\textbf{89.9} \\
        \midrule
        \multirow{3}{*}{\rotatebox{90}{\bf 72B}}
        &UI-TARS-72B & 94.9 & 82.5 & 89.7 & 88.6 & 88.7 & 85.0 & 88.4 \\
        &Aguvis-72B & 94.5 & 85.2 & 95.4 & 77.9 & 91.3 & 85.9 & 89.2 \\
        &UGround-V1-72B & 94.1 & 83.4 & 94.9 & 85.7 & 90.4 & 87.9 & \textbf{89.4}\\
        \bottomrule
    \end{tabular}
    }
    }
    \label{tab:ssv1}
\end{table*}
\begin{table*}[!t]
    \centering
    \footnotesize
    \caption{Inference and training time (85k training set) comparison between \method{}-3B and GUI-Actor-3B.}
    \vspace{-1em}
    \begin{tabular}{lcc}
    \toprule
    & \textbf{GUI-Actor-3B} & \textbf{\method{}-3B} \\ 
    \midrule
    ScreenSpot-v2 (s/iteration)  &0.73  & 0.76 \\
    ScreenSpot-Pro (s/iteration) &1.72  & 1.76 \\
    OSWorld-G (s/iteration)      &1.15  &1.23  \\
    \midrule
    Training time (A100 GPU-hours) & $\sim260$ & $\sim192$\\
    \bottomrule
\end{tabular}    
    \label{tab:time}
\end{table*}
\begin{table*}[!t]
    \centering
    \footnotesize
    \caption{Ablation results of \method{}-lite and GUI-Actor~\cite{wu2025gui} using the grounding verifier of GUI-Actor on ScreenSpot-v1 and ScreenSpot-v2.}
    \vspace{-1em}
    \resizebox{0.9\linewidth}{!}{
    \begin{tabular}{clccccccc}
        \toprule
        & \multirow{3}{*}{\textbf{Model}} 
          & \multicolumn{2}{c}{\textbf{Mobile}}
          & \multicolumn{2}{c}{\textbf{Desktop}}
          & \multicolumn{2}{c}{\textbf{Web}}
          & \multirow{2.5}{*}{\textbf{Avg.}} \\
        \cmidrule(lr){3-4}\cmidrule(lr){5-6}\cmidrule(lr){7-8}
        & & Text & Icon & Text & Icon & Text & Icon &  \\
        \midrule
        &GUI-Actor-3B (Qwen2.5-VL) + Verifier & 95.6&84.7&93.8&83.6&93.0&83.0&89.5$_{(+1.1)}$ \\
        &GUI-Actor-7B (Qwen2.5-VL) + Verifier & 96.0&85.6&96.4&82.9&91.7&83.0&\textbf{89.9}$_{(+0.0)}$\\
        &\textbf{\method{}-3B-lite} + Verifier & 97.1&83.8&96.4&83.6&91.3&84.0&\textbf{89.9}$_{(+1.1)}$ \\
        \midrule
        &GUI-Actor-3B (Qwen2.5-VL) + Verifier & 98.3 & 85.3 & 96.9 & 87.9 & 95.3 & 86.7 & 92.4$_{(+1.4)}$ \\
        &GUI-Actor-7B (Qwen2.5-VL) + Verifier & 96.9 & 89.6 & 97.4 & 86.4 & 95.7 & 84.7 & 92.5$_{(+0.4)}$ \\
        &\textbf{\method{}-3B-lite} + Verifier & 99.0&87.2&99.0&89.3&96.6&83.3&\textbf{93.0}$_{(+1.5)}$ \\
        \bottomrule
    \end{tabular}
    }
    \label{tab:results_with_verifier}
\end{table*}
\begin{table*}[!t]
    \centering
    \footnotesize
    \caption{Ablation results on ScreenSpot-v2, ScreenSpot-Pro and OSWorld-G of GUI grounding methods fine-tuned with the entire training set (254k elements). All variants are trained with the weighted patch-wise label introduced in \cref{sec:label} except $^*$.}
    \vspace{-1em}
    \resizebox{1\linewidth}{!}{
    \begin{tabular}{p{300pt}ccccccc}
        \toprule
        \multirow{2}{*}{\textbf{Model}} & \multicolumn{3}{c@{\hspace{1em}}}{\textbf{ScreenSpot-v2}} & \multicolumn{3}{c@{\hspace{1em}}}{\textbf{ScreenSpot-Pro}} & \multicolumn{1}{c@{\hspace{1em}}}{\textbf{OSWorld-G}}\\
        \cmidrule(lr){2-4}\cmidrule(lr){5-7}\cmidrule(lr){8-8}
        & Text & Icon & \textbf{Avg.}& Text & Icon & \textbf{Avg.} & \textbf{Avg.} \\
        \midrule
        GUI-Actor-3B$^*$ w/o weighted patch-wise label &94.99&80.87&88.84 & 49.13&19.04&37.63&50.53 \\
        GUI-Actor-3B &95.13&83.94&90.25&54.04&24.50&42.76&55.14\\
        \midrule
        \rowcolor{gray!10}\multicolumn{8}{l}{\textit{Attention Grouding without $\texttt{<ANCHOR>}$}}\\
        Vanilla Attention Grounding (multi-head weighting uniformly) &95.40&81.41&89.31& 58.03&21.52&44.09&55.03\\
        Vanilla Attention Grounding (multi-head weighted with $\mathcal{Q}'=\mathcal{Q}$ in \cref{eq:unified_head_score}) & 95.40&81.23&89.23 &55.07&23.84&43.14&54.14 \\
        \midrule
        \rowcolor{gray!10}\multicolumn{8}{l}{\textit{Simplified Attention Grouding with $\texttt{<ANCHOR>}$}}\\
        $\texttt{<ANCHOR>}$-based multi-head weighting with $\mathcal{Q}'=\{\texttt{<ANCHOR>}\}$ in \cref{eq:unified_head_score}&96.52&82.85&90.57&59.77&21.69&45.22&54.79\\
        GUI-AIMA-3B-lite w/ stop-grad on $w$&96.38&85.56&\textbf{91.67}&58.34&25.33&45.73&56.38 \\
        GUI-AIMA-3B-lite &97.21&84.12&91.51& 62.03&29.97&\textbf{49.78}&\textbf{58.33} \\
        \bottomrule
      \end{tabular}}
    \label{tab:ablation_all}
\end{table*}
\begin{figure}
\centering
\includegraphics[width=0.8\linewidth]{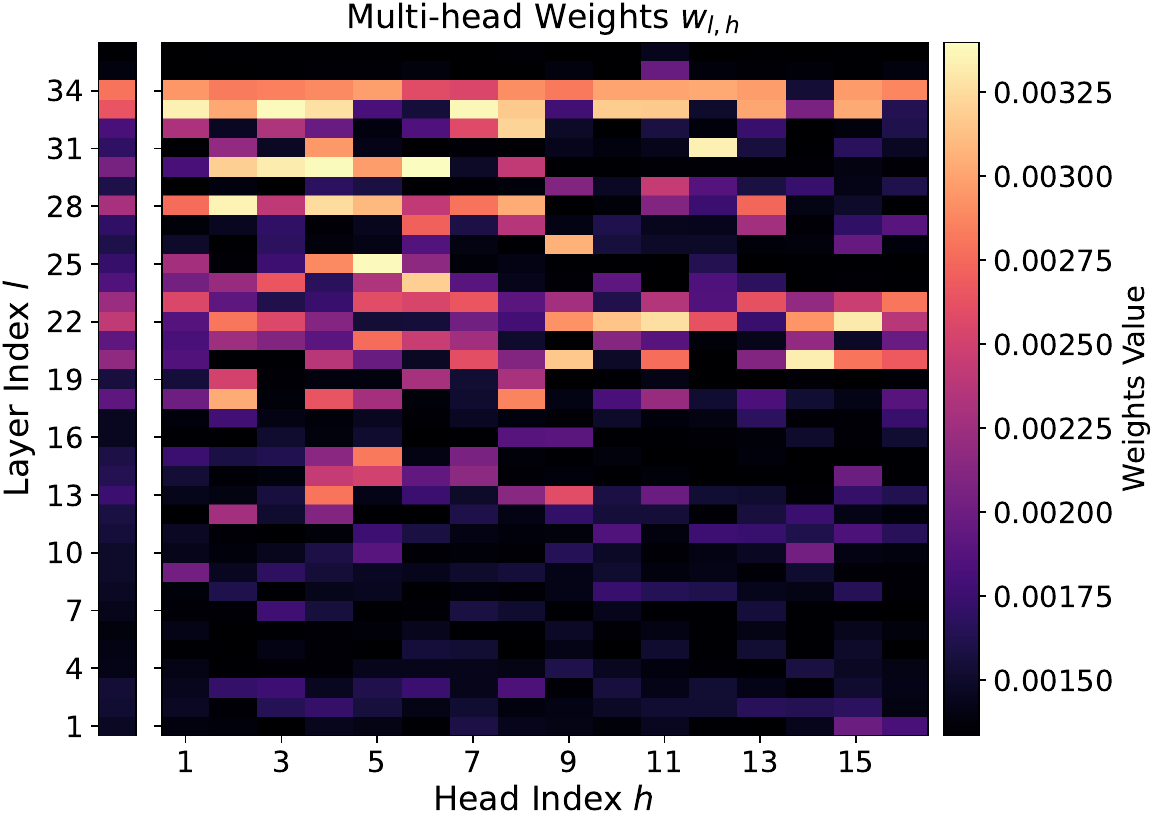}
\caption{Visualization of multi-head weights $w_{l,h}$ averaged over the ScreenSpot-Pro, ScreenSpot-v2 and OS-World-G. \textbf{Left:} layer-wise mean, i.e., for layer $l$: $\frac{1}{H}\sum_{h=1}^{H}w_{l,h}$. \textbf{Right:} per-layer, per-head weights $w_{l,h}$.
}
\label{fig:contribution_head}
\end{figure}
\section{Attention Head Contribution in GUI Grounding}
\label{app:attn_contribution}
In \cref{fig:contribution_head}, we visualize the head weights $w_{l,h}$ used during inference of \method{}. Overall, higher weights concentrate in mid-to-late layers (e.g., 22, 28, 33, 34), indicating that deep layers' representations contribute more to grounding. Mid layers contain highly dominant heads (e.g., head 25-5), whereas late layers (e.g., layer 34) tend to exhibit a broader set of moderately high-weight heads.

To quantify the effect of head selection, we perform inference-time head pruning based on $w_{l,h}$. Preserving only the top 30\% heads produces nearly unchanged accuracy (91.0\% on ScreenSpot-v2 and 53.5\% on ScreenSpot-Pro). In contrast, preserving only the bottom 30\% heads causes a substantial drop (76.6\% on ScreenSpot-v2 and 38.1\% on ScreenSpot-Pro). These results confirm that $w_{l,h}$ serves as an effective indicator of grounding-relevant attention heads. Additional visualizations of $w_{l,h}$ are provided in the supplementary material.

\section{Analysis for Visual-sink Query Token} \label{app:qs_analysis}
From ablation results in \cref{sec:ablation}, the global pattern indicated from the visual-sink query token computed $\mathcal{Q}_s$ from hidden states achieved better performance, supporting our selection that complies with the visual-query pattern from hidden states for weighting attention grounding.
\begin{figure}
    \centering
    \includegraphics[width=0.8\linewidth]{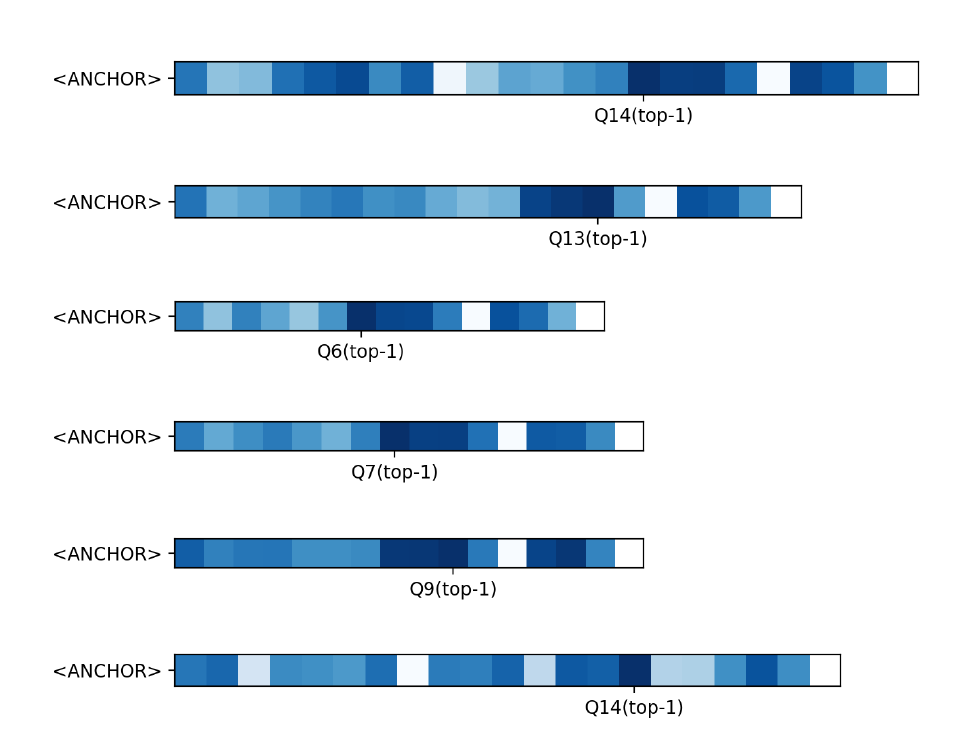}
    \caption{Cosine-similarity between multi-head attention vector computed between $\texttt{<ANCHOR>}$ and each text token. The top-1 visual-query token selected by \cref{eq:find_sink} consistently has the largest value with the $\texttt{<ANCHOR>}$ token.}
    \label{fig:visual_token}
\end{figure}
In \cref{fig:visual_token}, we compute the similarity values between the multi-head weight vector computed from $\texttt{<ANCHOR>}$ token as the $\mathcal{Q}'=\{\texttt{<ANCHOR>}\}$ variant in \cref{eq:unified_head_score} and computed from each query token $q_i$ as $w_{l,h}=\sum_{v_i \in \mathcal{V}}\,\mathbf{A}^{l,h}_{q_i,v_i}$ from the trained \method{} model. The top-1 visual-sink query token consistently has the most similar multi-head weight vector with it computed from $\texttt{<ANCHOR>}$, supporting our design that computing $w_{l,h}$ from the visual-sink query token instead of $\texttt{<ANCHOR>}$ token, as it can work similarly with $\texttt{<ANCHOR>}$ in computing multi-head weight vector but avoiding the bias from random initialized $\texttt{<ANCHOR>}$ token. 
\section{Baselines Details}\label{app:baseline}
Most baselines' results are taken from the original papers, except for GUI-G$^2$-3B, which is not reported in the source. Evaluation results of baselines for UI-Vision and MMBenchGUI-L2 in \cref{tab:ui_mm_v2} are taken from the results reported in papers of UI-Venus-1.5~\cite{gao2026ui} and GroundNext~\cite{feizi2025grounding}.
\section{Extra details of $\texttt{<ANCHOR>}$ Implementations} \label{app:anchor}
In \cref{sec:pre}, we abbreviate the implementation of $[\mathcal{V},\mathcal{Q},\texttt{<ANCHOR\_START>},\texttt{<ANCHOR>},\\\texttt{<ANCHOR\_END>}]$ as $[\mathcal{V},\mathcal{Q},\texttt{<ANCHOR>}]$ for brevity. For the single-area prediction setting in GUI grounding, we explore expanding a single $\texttt{<ANCHOR>}$ to multiple $\texttt{<ANCHOR>}$ tokens as $[\mathcal{V},\mathcal{Q},\texttt{<ANCHOR\_0>}, \texttt{<ANCHOR\_1>},\ldots,\texttt{<ANCHOR\_N>}]$ (abbreviate start and end tokens). However, it turns out to bring no performance gains and merely adds redundancy for the single-region grounding tasks. We will explore the multi-region grounding tasks with a disentangled $\texttt{<ANCHOR\_n>}$ token for each grounding region as future work.

\section{GUI Grounding Examples of \method{}}
\label{app:example}
\begin{figure}[t]
  \centering
  \begin{subfigure}[t]{0.37\linewidth}
    \centering
    \includegraphics[width=\linewidth]{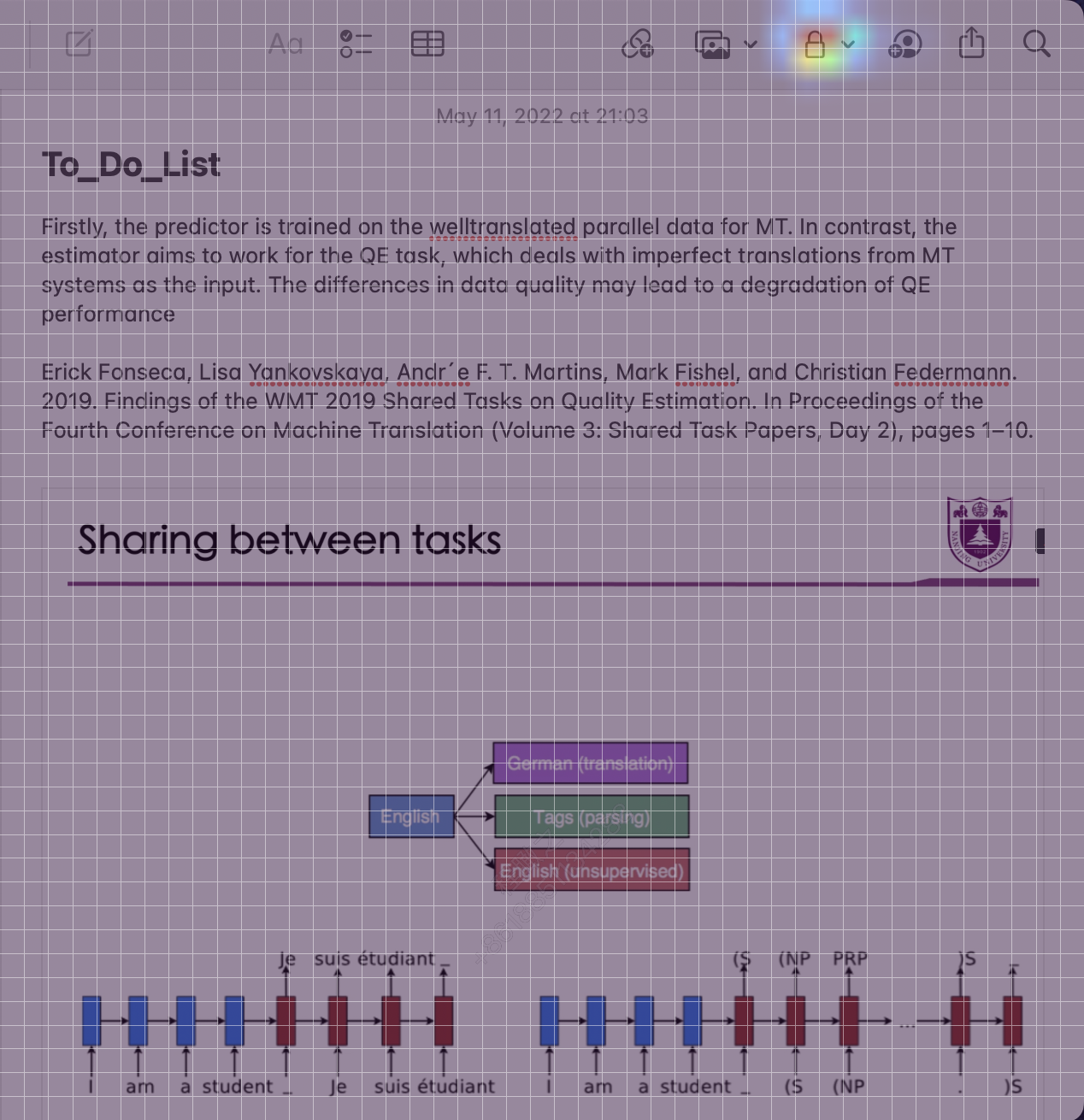} 
    \caption{Lock the memo.}
    \label{fig:v2_1}
  \end{subfigure}\hfill
  \begin{subfigure}[t]{0.37\linewidth}
    \centering
    \includegraphics[width=\linewidth]{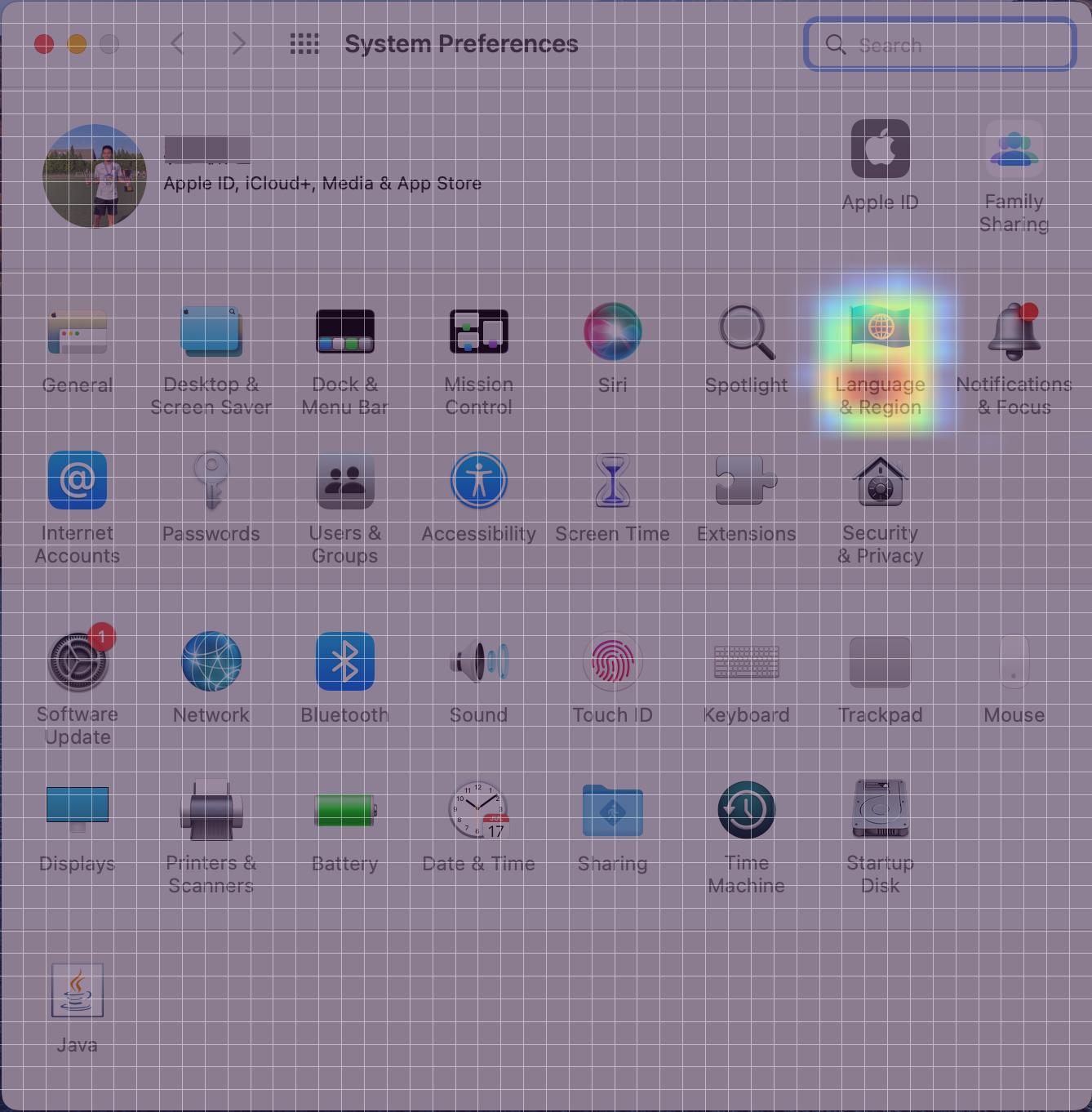}
    \caption{View the language \& region settings.}
    \label{fig:v2_2}
  \end{subfigure}\hfill
  \begin{subfigure}[t]{0.25\linewidth}
    \centering
    \includegraphics[width=\linewidth]{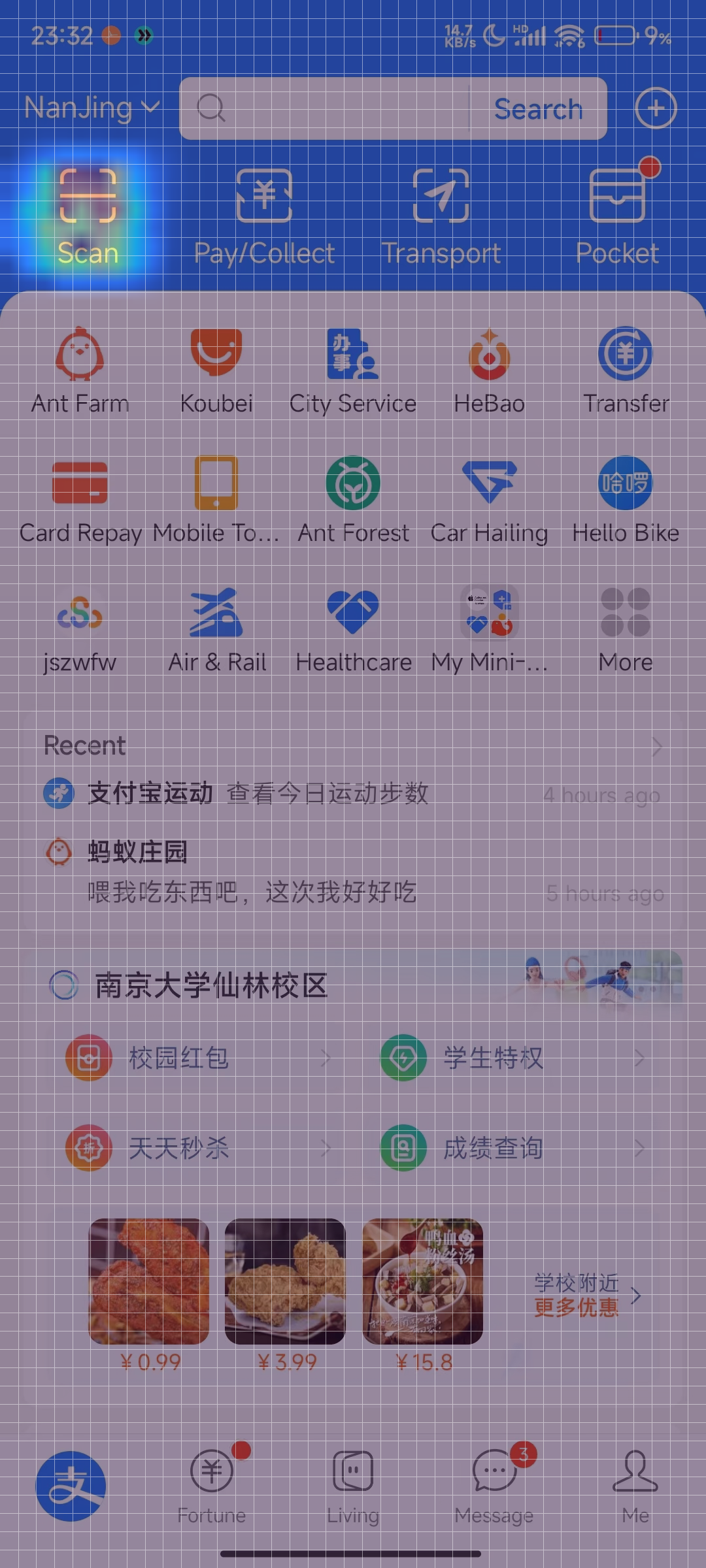}
    \caption{Scan QR code.}
    \label{fig:v2_3}
  \end{subfigure}
  \begin{subfigure}[t]{0.54\linewidth}
    \centering
    \includegraphics[width=\linewidth]{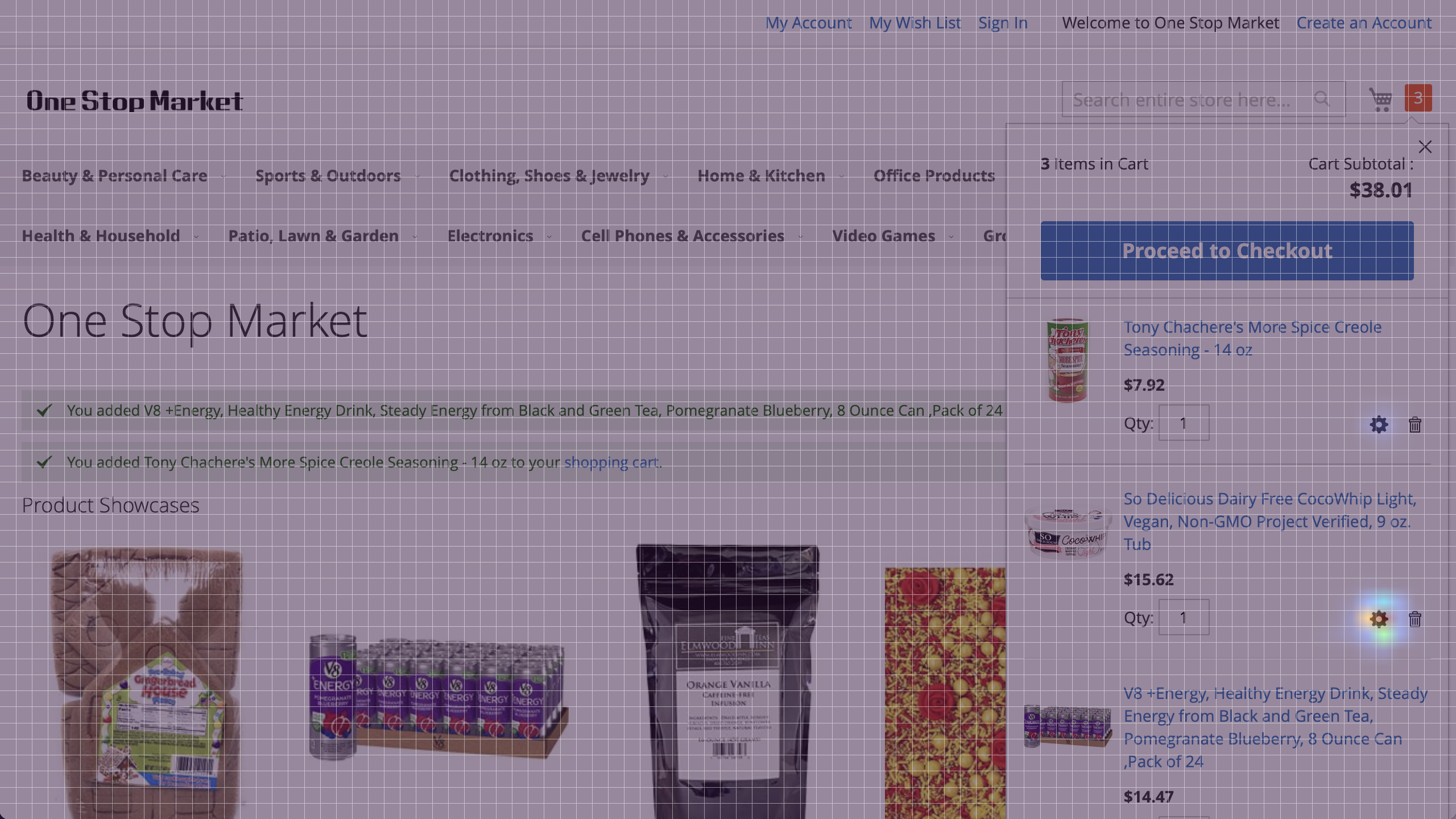}
    \caption{View settings of cocowhip light.}
    \label{fig:v2_4}
  \end{subfigure}\hfill
  \begin{subfigure}[t]{0.44\linewidth}
    \centering
    \includegraphics[width=\linewidth]{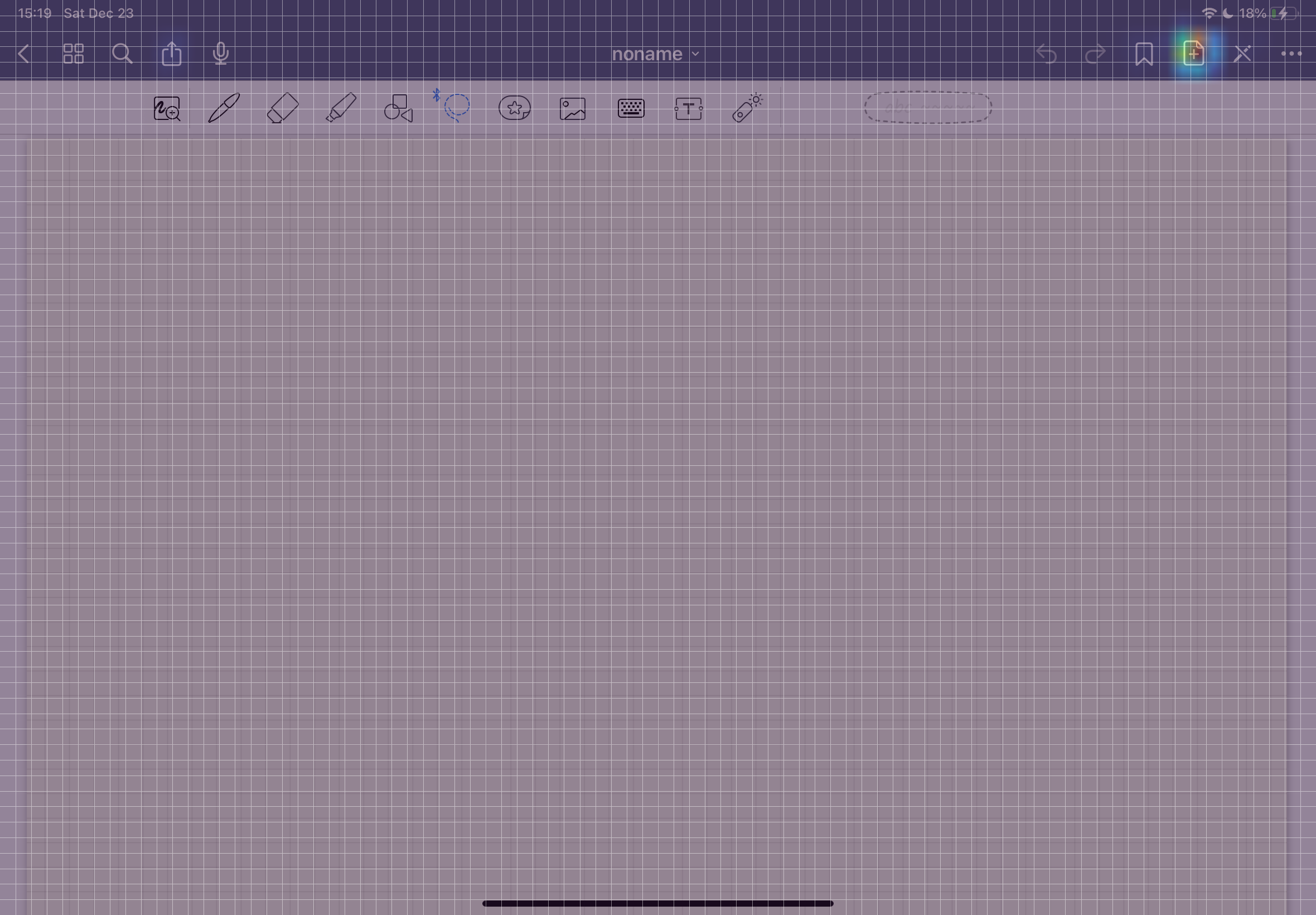}
    \caption{Add a new page.}
    \label{fig:v2_5}
  \end{subfigure}
  \caption{Visualization examples of \method{}'s grounding results on ScreenSpot-v2.}
  \label{fig:example_v2}
\end{figure}

\begin{figure}[!htbp]
\centering
\begin{subfigure}[b]{0.9\linewidth}
    \includegraphics[width=\linewidth]{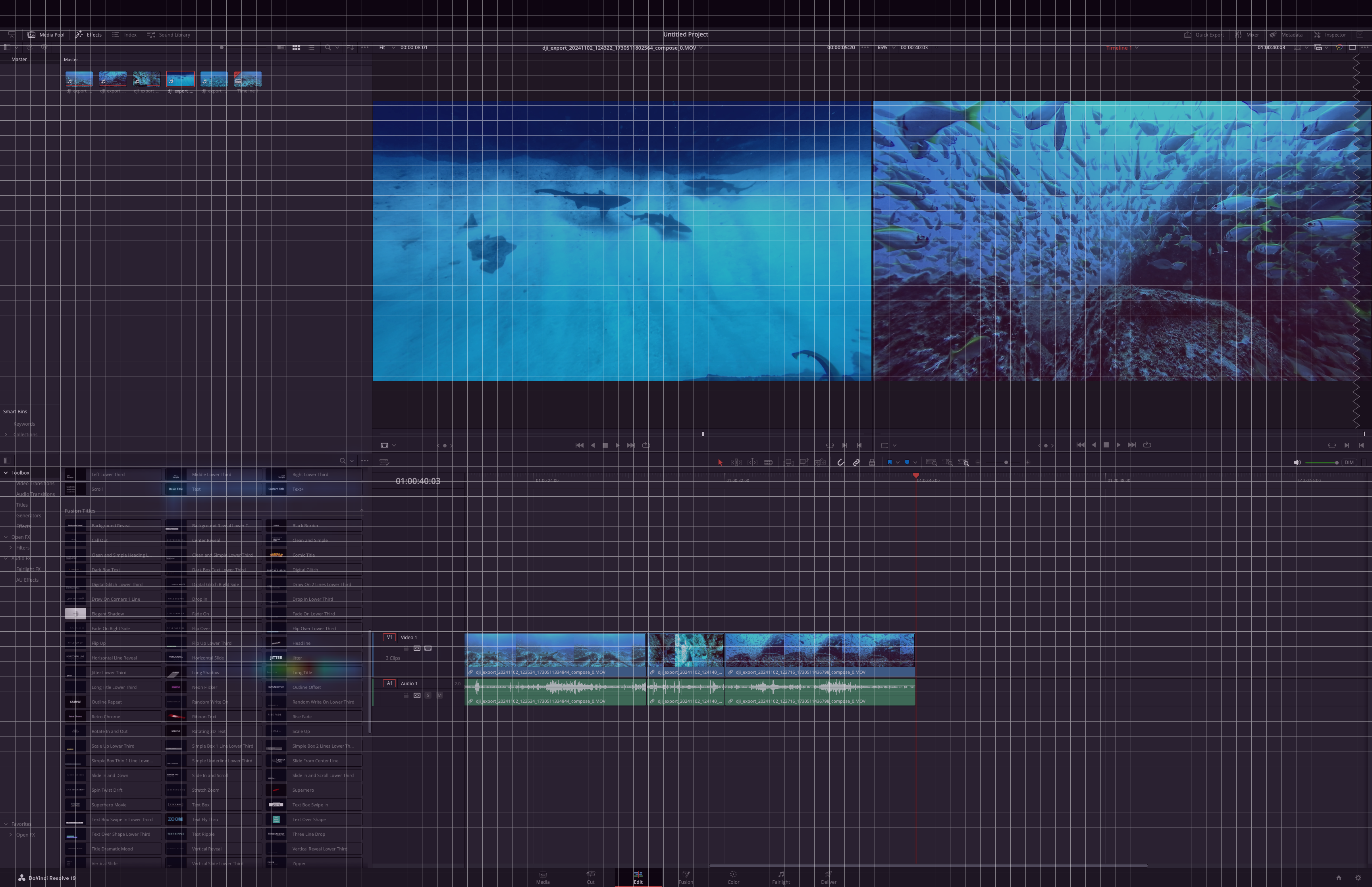}
    \caption{Add long title.}
\end{subfigure}

\vspace{2pt}

\begin{subfigure}[b]{0.9\linewidth}
    \includegraphics[width=\linewidth]{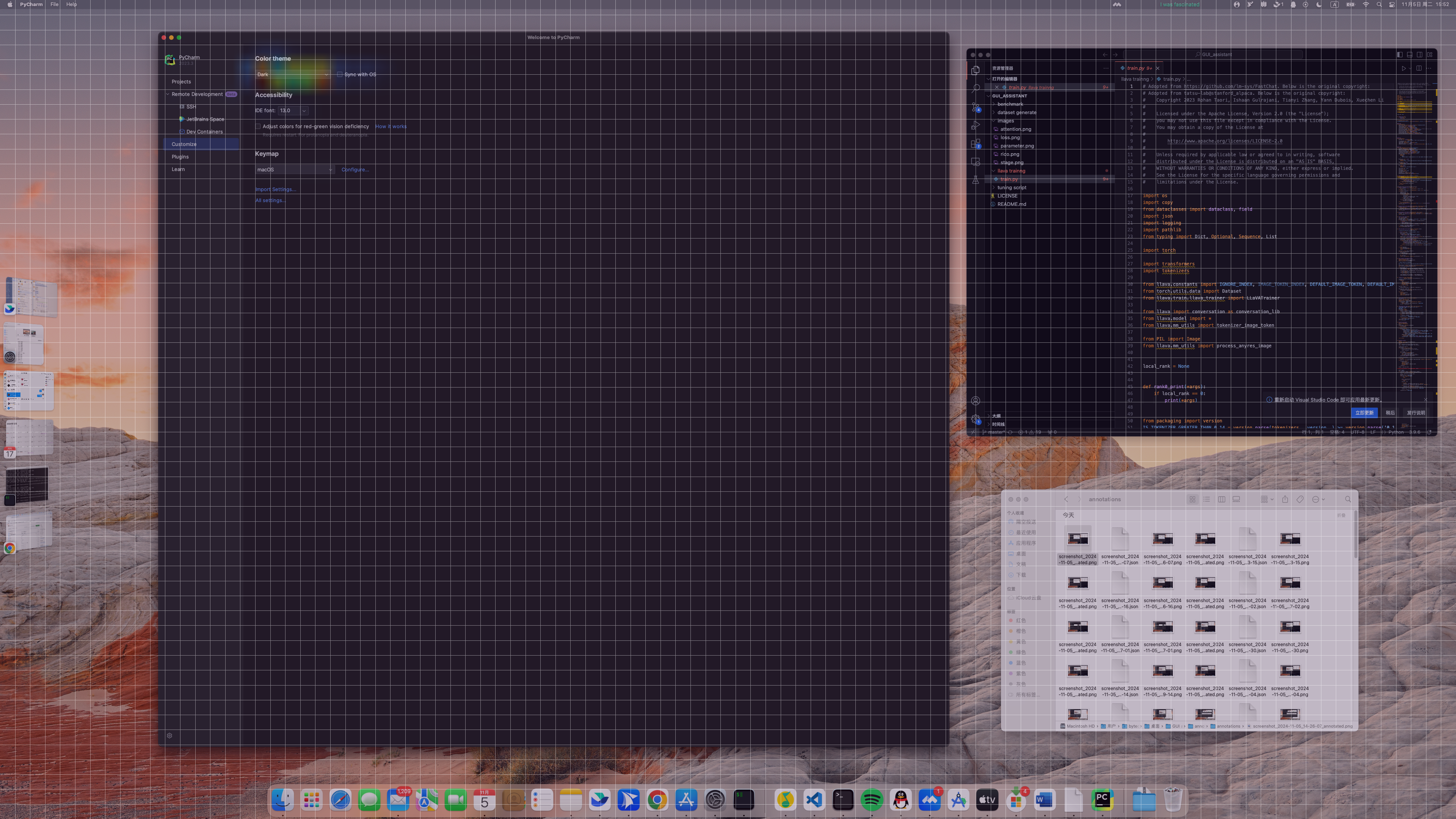}
    \caption{Change color theme in PyCharm.}
\end{subfigure}
\caption{Visualization examples of \method{}'s grounding results on ScreenSpot-Pro.}
\label{fig:example_pro1}
\end{figure}

\begin{figure}[!htbp]
\centering
\begin{subfigure}[b]{0.9\linewidth}
    \includegraphics[width=\linewidth]{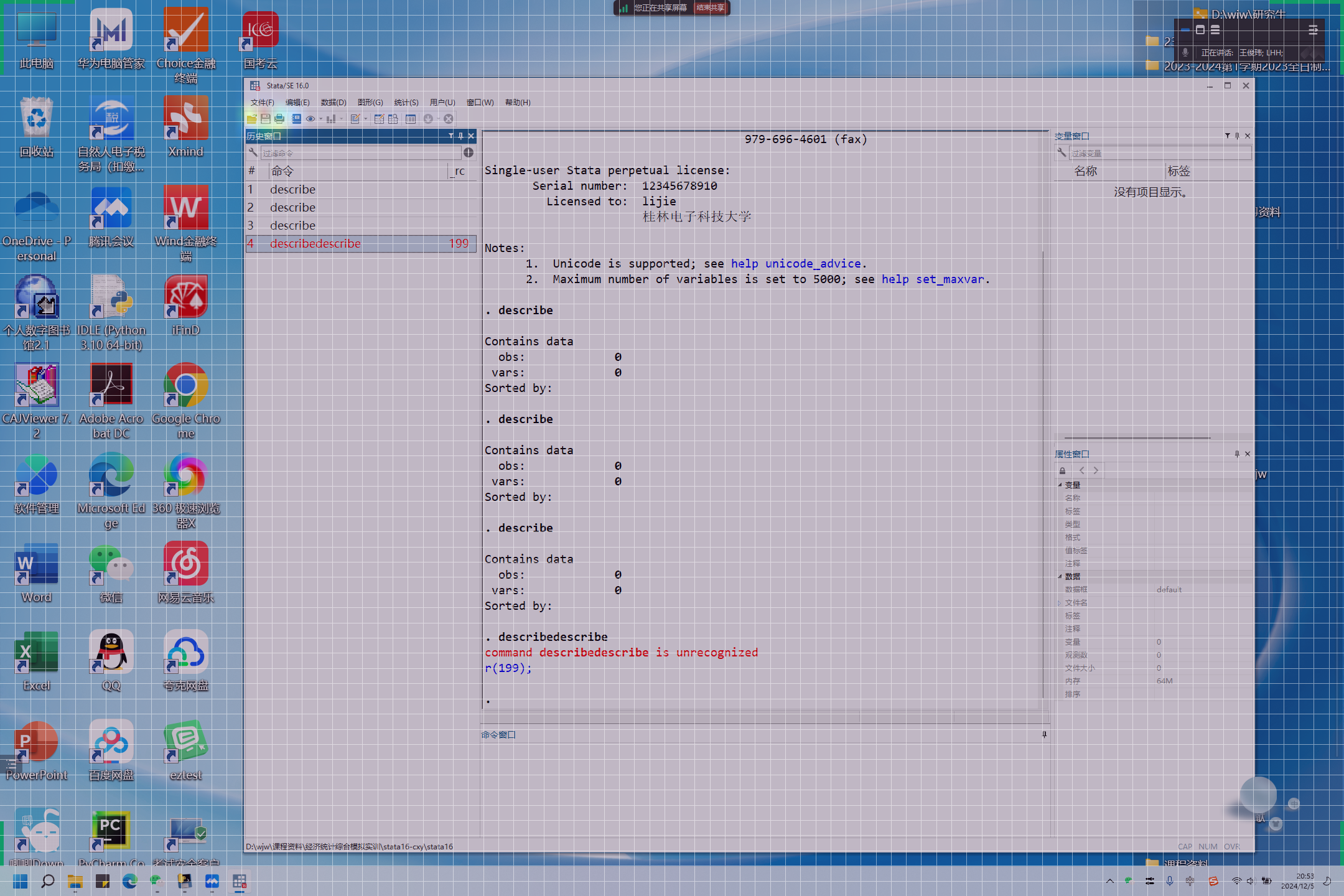}
    \caption{Save a file.}
\end{subfigure}

\vspace{2pt}

\begin{subfigure}[b]{0.9\linewidth}
    \includegraphics[width=\linewidth]{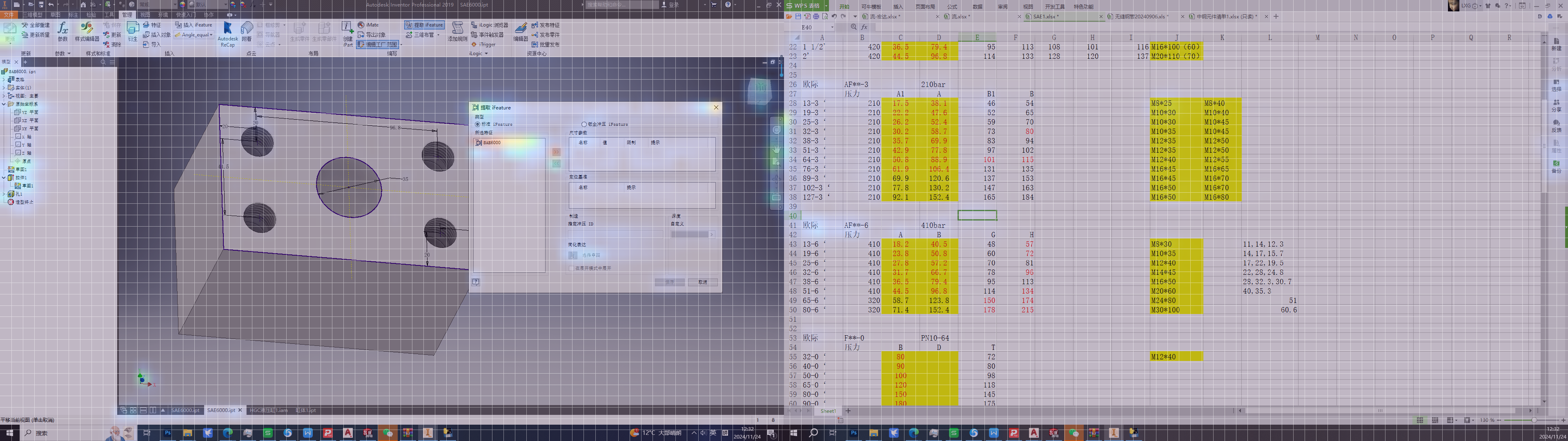}
    \caption{Select the only feature.}
\end{subfigure}

\caption{Visualization examples of \method{}'s grounding results on ScreenSpot-Pro.}
\label{fig:example_pro2}
\end{figure}

We provide the visualizations of \method{}'s multi-head attention grounding results on ScreenSpot-v2 and ScreenSpot-Pro as follows in \cref{fig:example_v2}, \cref{fig:example_pro1,fig:example_pro2}.

\section{Failure Cases}
\label{app:failure}
\begin{figure}[t]
    \centering
    \includegraphics[width=\linewidth]{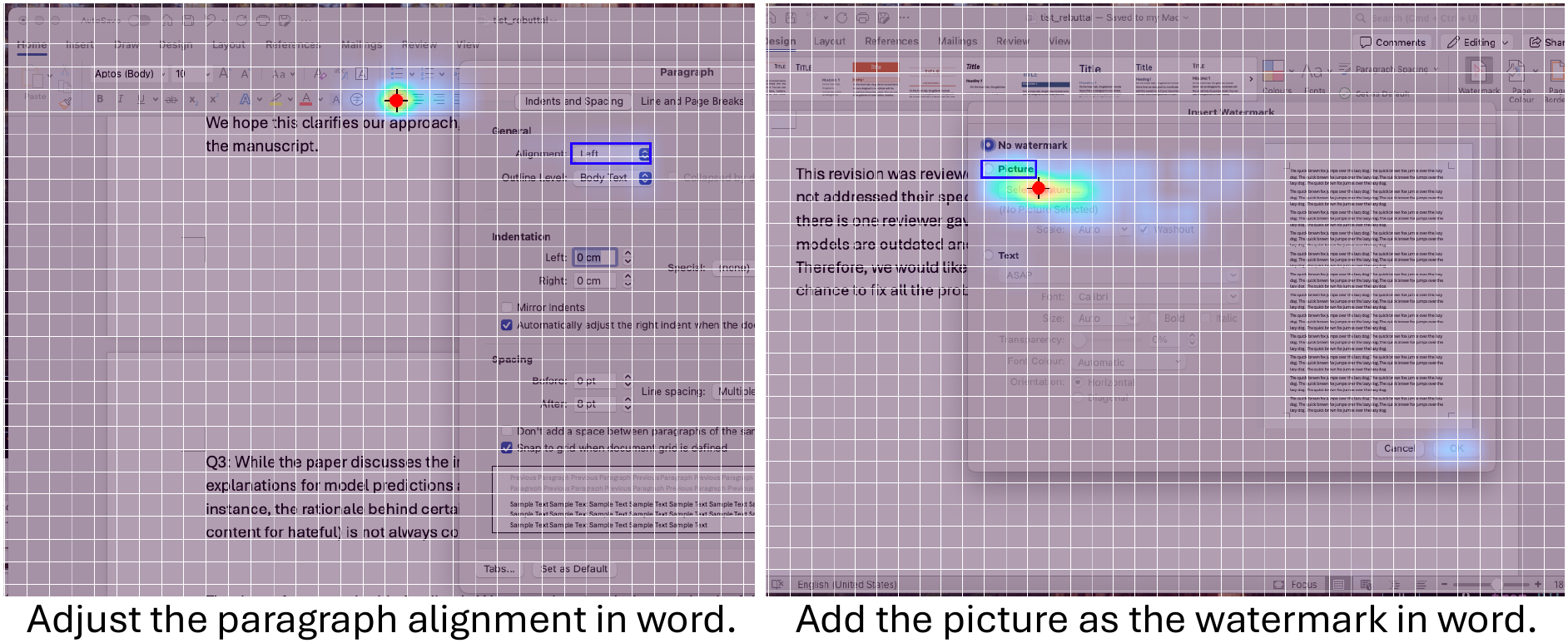}
    \caption{Failure cases of \method{} in Word on macOS.}
    \label{fig:failure_word}
\end{figure}

In \cref{fig:two_step} of the main paper, we show a failure case where \method{} finds the dense tool area but fails to recognize the true icon due to low granularity. In \cref{fig:failure_word}, we show additional failure cases in Word on macOS: (1) \method{} clicks a non-clickable button with the same function, ignoring the ``Paragraph'' window; (2) incorrect action order: it should first click ``Picture'' before ``Select Picture''.

\end{document}